\pdfoutput=1
\documentclass[11pt]{article}
\usepackage[]{acl}
\usepackage{times}
\usepackage{colortbl} 
\usepackage{latexsym}
\usepackage{graphicx}
\usepackage[T1]{fontenc}
\usepackage[utf8]{inputenc}
\usepackage{microtype}
\usepackage{inconsolata}
\usepackage[addedmarkup=uwave,defaultcolor=red]{changes}
\usepackage{url}
\usepackage{tcolorbox}
\usepackage{graphicx}
\usepackage{makecell}
\usepackage{framed}
\usepackage{amsmath}
\usepackage{enumitem}
\usepackage{pifont}
\usepackage{wrapfig}
\usepackage{booktabs}
\usepackage{amsfonts}       
\usepackage{nicefrac}       
\usepackage{microtype}      
\usepackage{xspace}
\usepackage{makecell}
\usepackage{multirow}
\usepackage[capitalize]{cleveref}
\usepackage{scalerel}
\usepackage[misc]{ifsym}
\usepackage{float}
\usepackage{stfloats}
\usepackage{subcaption} 
\usepackage{ulem}
\usepackage{fontawesome5}
\usepackage[outline]{contour}
\usepackage{amssymb}
\usepackage{dsfont}
\usepackage{algorithm}
\usepackage{algpseudocode}
\usepackage{amsmath}
\usepackage{times}
\usepackage{latexsym}
\usepackage{tabularx}
\usepackage{float}       
\usepackage{placeins} 
\usepackage{pifont}
\usepackage{todonotes}
\let\Letter\relax
\usepackage{marvosym}

\definecolor{darkgreen}{rgb}{0.09, 0.46, 0.16}
\definecolor{darkred}{rgb}{0.72, 0.13, 0.07}
\definecolor{iccvblue}{rgb}{0.21,0.49,0.74}
\definecolor{sbblue}{HTML}{5273AD}
\definecolor{sborange}{HTML}{D78357}
\definecolor{sbgreen}{HTML}{5fA86C}
\definecolor{sbred}{HTML}{BD4C53}

\newcounter{takeaway}
\newcounter{conclusion}
\newcounter{greentakeaway}
\definecolor{darkgreen}{RGB}{0,160,0}
\makeatletter
\DeclareRobustCommand\onedot{\futurelet\@let@token\@onedot}
\def\@onedot{\ifx\@let@token.\else.\null\fi\xspace}

\def\ie{\emph{i.e}\onedot}

\makeatother

\makeatletter
\renewcommand{\paragraph}{%
  \@startsection{paragraph}{4}%
  {\z@}{0ex \@plus 0ex \@minus 0ex}{-1em}%
  {\normalfont\normalsize\bfseries}%
}
\makeatother

\definecolor{shadecolor}{rgb}{0.92,0.92,0.92}
\definecolor{bblue}{HTML}{0000fd}
\definecolor{rred}{HTML}{fe0000}
\definecolor{ggreen}{HTML}{018201}
\definecolor{redd}{HTML}{680020}
\definecolor{reddd}{HTML}{b71a36}
\definecolor{redddd}{HTML}{ebb9a7}

\crefname{figure}{Fig.}{Figs.}
\crefname{table}{Tab.}{Tabs.}
\crefname{section}{Sec.}{Secs.}

\title{Understanding and Leveraging the Expert Specialization of Context Faithfulness in Mixture-of-Experts LLMs}

\author{Jun Bai\textsuperscript{1}\quad
        Minghao Tong\textsuperscript{1,2}\quad
        Yang Liu\textsuperscript{1}\quad
        Zixia Jia\textsuperscript{1,\Letter}\quad
        Zilong Zheng\textsuperscript{1,\Letter}\\
  \textsuperscript{1} State Key Laboratory of General Artificial Intelligence, BIGAI\\
  \textsuperscript{2} School of Computer Science, Wuhan University\\
  \{baijun, liuyang, jiazixia, zlzheng\}@bigai.ai, tongminghao@whu.edu.cn
}

\begin{document}
\maketitle

\begin{abstract}
Context faithfulness is essential for reliable reasoning in context-dependent scenarios. However, large language models often struggle to ground their outputs in the provided context, resulting in irrelevant responses.
Inspired by the emergent expert specialization observed in mixture-of-experts architectures, this work investigates whether certain experts exhibit specialization in context utilization—offering a potential pathway toward targeted optimization for improved context faithfulness.
To explore this, we propose Router Lens, a method that accurately identifies context-faithful experts. Our analysis reveals that these experts progressively amplify attention to relevant contextual information, thereby enhancing context grounding.
Building on this insight, we introduce Context-faithful Expert Fine-Tuning (CEFT), a lightweight optimization approach that selectively fine-tunes context-faithful experts.
Experiments across a wide range of benchmarks and models demonstrate that CEFT matches or surpasses the performance of full fine-tuning while being significantly more efficient\footnote{Our code is publicly available at \url{https://github.com/bigai-nlco/RouterLens}.}.
\end{abstract}

\section{Introduction}

Faithfulness to the provided context is essential for ensuring the reliability and coherence of responses in many context-dependent scenarios, such as long sequence processing \cite{DBLP:conf/acl/LiWZZ24,DBLP:journals/corr/abs-2502-18890}, In-Context Learning (ICL) \cite{DBLP:conf/acl/0003LBLWLR25,DBLP:conf/iclr/QiYJ0LZ0Z25}, and Retrieval-Augmented Generation (RAG)
\cite{DBLP:conf/naacl/ShiHLTZY24,DBLP:conf/iclr/SunZ0XZYSL25}.
Despite their remarkable fluency, Large Language Models (LLMs) often generate outputs that are only loosely grounded in the given context or, in more concerning instances, hallucinate information not supported by it
\cite{DBLP:conf/emnlp/ZhouZPC23,DBLP:conf/emnlp/ChuangQHKKG24,DBLP:journals/tois/HuangYMZFWCPFQL25}. 

\begin{figure}[t]
\centering
\includegraphics[width=\columnwidth]{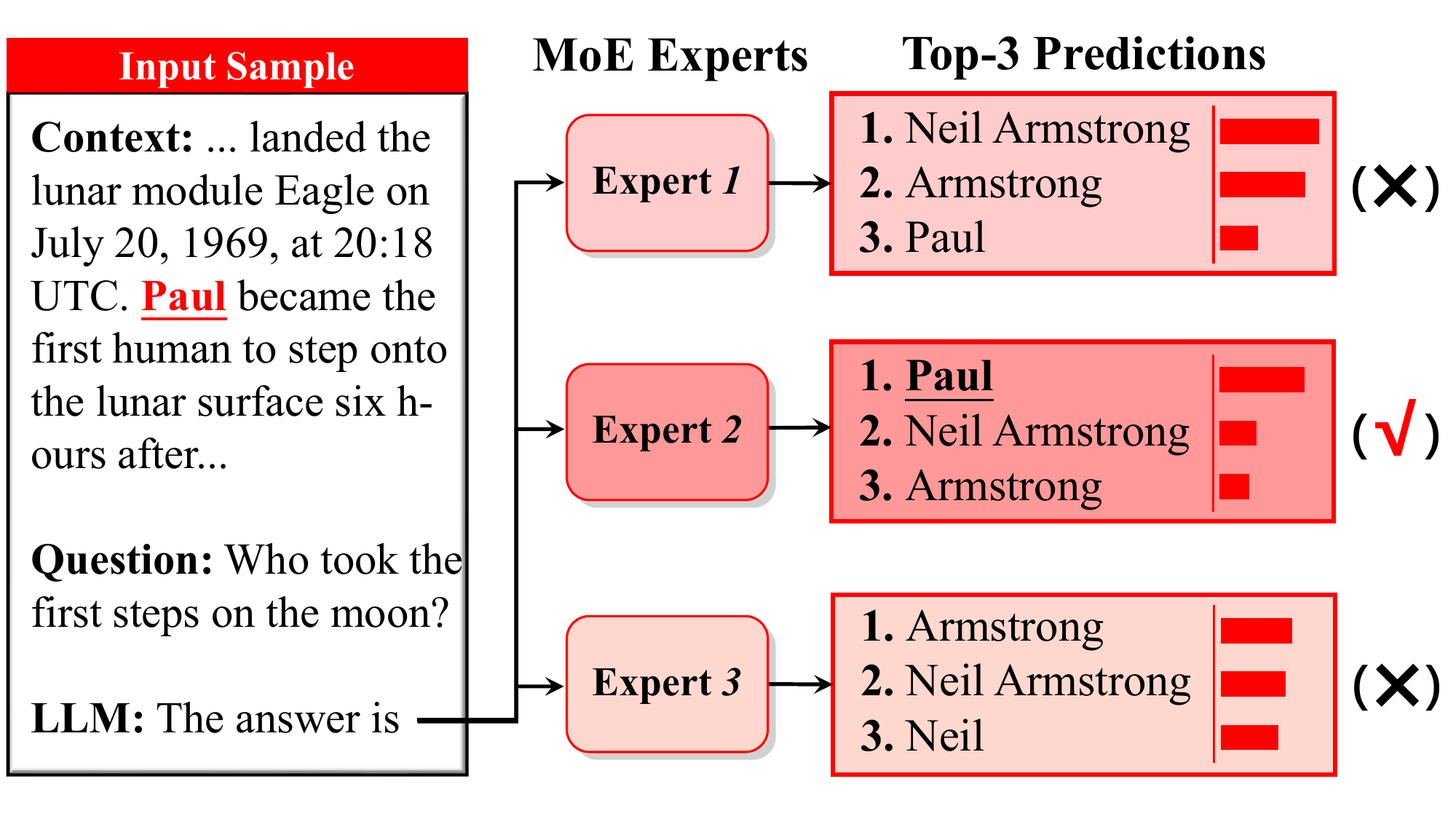}
\caption{A case from NQ-Swap \cite{DBLP:conf/emnlp/LongprePCRD021/nq-swap} where MoE experts exhibit different tendencies of context faithfulness (answer is \textcolor{red!65!black}{\underline{underlined in red}}).}
\label{fig:intro_case}
\end{figure}

Prior work has proposed sophisticated prompting \cite{DBLP:conf/emnlp/ZhouZPC23}, decoding \cite{DBLP:conf/naacl/ShiHLTZY24}, and alignment \cite{DBLP:journals/corr/abs-2412-15280/confliqa} techniques to enhance context faithfulness—the ability to accurately attend to and integrate relevant contextual information during response generation.
More recently, expert specialization in Mixture-of-Experts (MoE) models has emerged as a promising direction \cite{DBLP:conf/emnlp/WangCDXLW24/esft}, potentially enabling more targeted optimization of model capacity for context utilization.
Previous studies have shown that MoE experts tend to specialize in processing different aspects of input. The router network typically activates distinct experts when handling tokens from diverse tasks \cite{DBLP:journals/corr/abs-2401-04088/mixtral}, domains \cite{DBLP:journals/corr/abs-2409-02060/olmoe}, and syntactic units \cite{DBLP:conf/coling/AntoineBL25}.
Building on these observations, we identify that experts in MoE models exhibit varying degrees of context faithfulness, as illustrated in Figure~\ref{fig:intro_case}.
This raises an intriguing question: \textit{Do context-faithful experts exist within MoE models?}

In this work, we present a systematic analysis of expert specialization in MoE LLMs, with a particular focus on their ability to leverage contextual information.
A common approach to identifying experts responsible for specific functionalities involves analyzing expert activation frequencies across layers~\cite{DBLP:journals/corr/abs-2401-04088/mixtral,DBLP:conf/emnlp/WangCDXLW24/esft}, under the assumption that the most frequently activated experts are those most relevant to the task.
However, this assumption is undermined by the load balancing constraint imposed during pretraining
\cite{DBLP:conf/acl/DaiDZXGCLZYWXLH24/DeepSeekMOE}, which enforces uniform expert usage and consequently limits the router network’s ability to select the most beneficial experts
\cite{DBLP:conf/acl/Dai0MZSCW22/router}.
As a result, experts identified through standard activation-based heuristics may not accurately reflect optimal specialization for context-dependent behavior.

To investigate the above limitations, we propose \textbf{Router Lens}, an effective method for eliciting aspect-relevant experts by fine-tuning only the router network on context-dependent tasks, while keeping all other model parameters fixed.
This enables the model to dynamically route inputs to experts that are more effectively aligned with contextual information.
Experts that are frequently activated in the updated model are then identified as context-faithful experts.
We show that tuning the router alone significantly improves performance on context-dependent tasks, providing strong evidence that certain experts are indeed specialized for context utilization.
Further analysis reveals that these experts progressively amplify attention to contextually salient signals and guide intermediate representations toward more accurate outputs.

Motivated by the specialization potential of context-faithful experts, we move beyond full fine-tuning and introduce \textbf{Context-faithful Expert Fine-Tuning (CEFT)}, a parameter-efficient strategy that fine-tunes only the experts most relevant to context utilization.
CEFT enhances the model’s ability to utilize contextual information while substantially reducing the number of trainable parameters.
We evaluate CEFT across a diverse set of context-dependent tasks and models. Experimental results show that CEFT not only matches, but often surpasses, the performance of full fine-tuning.
Our contributions can be summarized as: 
\begin{itemize}[leftmargin=*, topsep=1pt, noitemsep]
\setlength{\itemsep}{5pt} 
      \item  We propose \textbf{Router Lens}, a novel framework for identifying context-faithful experts in MoE models. Our analysis shows that these experts play a critical role in effective context utilization.

      \item We introduce \textbf{Context-faithful Expert Fine-Tuning (CEFT)}, an optimization strategy targeted at the context-faithful experts identified by Router Lens, achieving competitive or superior performance compared to fully fine-tuning.

      \item We conduct comprehensive experiments across multiple benchmarks and models, demonstrating the generalizability and effectiveness of our approach in enhancing contextual faithfulness. 
      
\end{itemize}

\section{Preliminaries}

\subsection{Context-dependent Tasks}

In context-dependent task, the correct prediction or output depends not only on the input query $q$ but also on an additional context $c$, which provides essential supporting information \cite{DBLP:journals/air/KaziKD23,DBLP:conf/kdd/FanDNWLYCL24,DBLP:journals/corr/abs-2412-15280/confliqa}. Formally, let $\mathcal{Q}$ be the space of queries, $\mathcal{C}$ be the space of possible contexts, and $\mathcal{Y}$ be the space of outputs. A context-dependent task is defined by a function:
\begin{equation}\label{equ:context_task}
f: \mathcal{Q} \times \mathcal{C} \rightarrow \mathcal{Y}, \quad \text{such that} \quad y = f(q, c)
\end{equation}
where $y \in \mathcal{Y}$ is the task output. 

\subsection{Mixture-of-Experts LLMs}

MoE is an architecture where the Feed-Forward Network (FFN) \cite{DBLP:conf/nips/VaswaniSPUJGKP17/transformer_first} are replaced by MoE modules to efficiently increase model capacity \cite{DBLP:conf/icml/XueZFNZZ024/openmoe}. Each MoE layer consists of $N_e$ parallel experts $e$ (sharing FFN structure). 
For each input token, a subset of $k$ experts is activated (\ie top-$k$) based on learned gating scores computed by a router network \cite{DBLP:conf/acl/Dai0MZSCW22/router}.

Let $u_t^{(\ell)}$ be the input of the $t$-th token at the $\ell$-th MoE layer. The output $h_t^{(\ell)}$ is computed as:
\begin{equation}
h_t^{(\ell)} = u_t^{(\ell)} + \sum_{i \in \mathcal{S}_t} g_{i,t}^{(\ell)} \cdot \mathrm{FFN}_i^{(\ell)}\left(u_t^{(\ell)}\right)
\end{equation}
where $\mathcal{S}_t$ is the set of top-$k$ experts for token $t$, and $g_{i,t}^{(\ell)}$ are the corresponding gating weights.

The gating weights are produced by the router network, parameterized by $\theta_r^{(\ell)}$, which computes a score vector and applies a softmax operation:
\begin{equation}
s_{t}^{(\ell)} = \mathrm{Router}^{(\ell)}(u_t^{(\ell)}; \theta_r^{(\ell)}) \in \mathbb{R}^{N_e}
\end{equation}
\begin{equation}
\tilde{s}_{i,t}^{(\ell)} = \frac{\exp\left(s_{i,t}^{(\ell)}\right)}{\sum_{j \in \mathcal{S}_t} \exp\left(s_{j,t}^{(\ell)}\right)}
\end{equation}
\begin{equation}
g_{i,t}^{(\ell)} =
\begin{cases}
\tilde{s}_{i,t}^{(\ell)}, & \text{if } i \in \mathcal{S}_t, \\
0, & \text{otherwise}.
\end{cases}
\end{equation}
Here, $\mathrm{Router}^{(\ell)}(\cdot; \theta_r^{(\ell)})$ denotes the router network in layer $\ell$. Only the experts with the top-$k$ gating scores contribute to the final output.

\section{Eliciting Context-faithful Experts}

\subsection{Router Lens}
To identify experts relevant to a specific capability, prior work often uses expert activation frequency as a proxy—experts that are activated more frequently are assumed to be more important \cite{DBLP:journals/corr/abs-2401-04088/mixtral,DBLP:conf/emnlp/WangCDXLW24/esft}.
However, in pretrained MoE models, the router network is typically optimized with a strong load balancing constraint to enforce uniform expert usage \cite{DBLP:journals/corr/abs-2408-15664}.
While this constraint improves training stability and computational efficiency, it can obscure the natural emergence of expert specialization for particular capabilities such as context utilization.
As a result, models may struggle to accurately identify and leverage the most beneficial experts for context-dependent tasks.

\begin{figure}[t]
\centering
\includegraphics[width=\columnwidth]{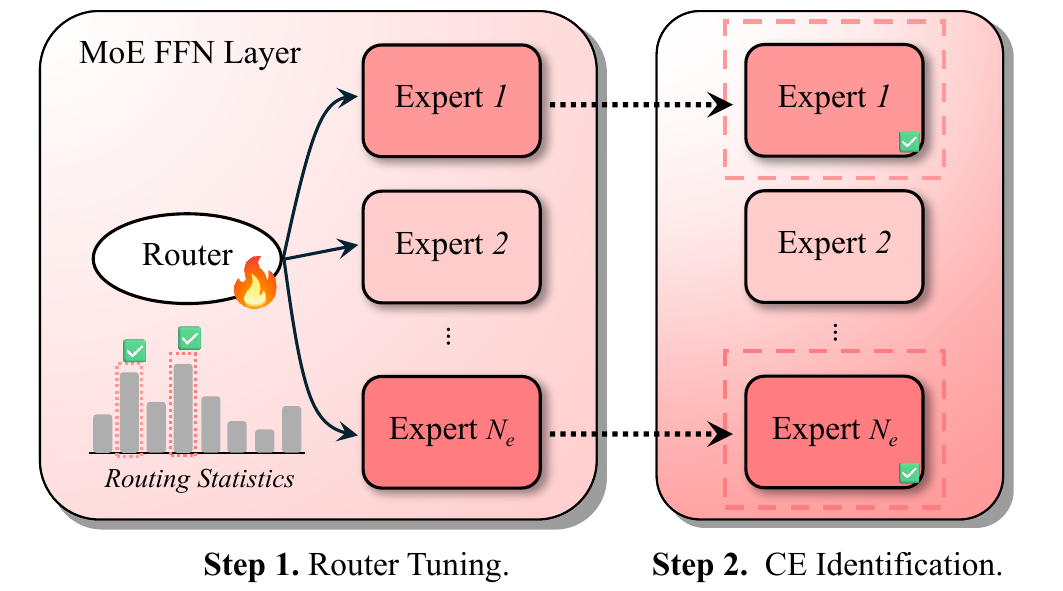}
\caption{Illustration of Router Lens, where the Context-faithful Experts (CE) in each layer are identified by the tuned router network.}
\label{fig:router_tuning}
\end{figure}

To address this limitation, we propose a novel expert elicitation method, \textbf{Router Lens}.
As shown in Figure \ref{fig:router_tuning}, it begins with a lightweight adaptation step—router tuning, which encourages the router network to relearn expert selection tailored to a specific context-dependent task.
Following this, we introduce a context-dependence ratio, computed using the tuned router network, as a principled metric for identifying context-faithful experts.

Specifically, let the MoE parameters be denoted as $\theta = \{\theta_r, \theta_o\}$, where $\theta_r$ are the parameters of all router networks across layers, and $\theta_o$ represents all remaining parameters of the model, including the experts, attention layers, embeddings, and layer norms, etc.
In router tuning, we freeze $\theta_o$ and update only the router parameters $\theta_r$ to minimize the context-dependent task loss. Formally, the optimization objective is:
\begin{equation}
\min_{\theta_r} \;\; \mathcal{L}_{\text{task}}\left(f(x; \theta_r, \theta_o)\right)
\quad \text{s.t. } \theta_o \text{ fixed},
\end{equation}
where $x$ is the model input, $f(\cdot)$ is the MoE model forward function, and $\mathcal{L}_{\text{task}}$ is the supervised loss specific to the context-dependent task.
In this way, gradient descent optimization is applied only with respect to $\theta_r$:
\begin{equation}
\theta_r \leftarrow \theta_r - \eta \cdot \nabla_{\theta_r} \mathcal{L}_{\text{task}}(f(x; \theta_r, \theta_o))
\end{equation}
where $\eta$ denotes the learning rate.

To identify the context-faithful experts, we define the \textbf{Context-dependence Ratio} $r_i^{(\ell)}$ for expert $e_i$ in the $\ell$-th layer, measuring the frequency for which expert $i$ is selected by the tuned router network across the training dataset. Formally:
\begin{equation}
r_i^{(\ell)} = \frac{1}{N_s} \sum_{j=1}^{N_s} \frac{1}{L_j} \sum_{t=1}^{L_j} \frac{\mathds{1}\left(g_{i,t}^{(\ell,j)} > 0\right)}{k}
\end{equation}
where $N_s$ is the total number of input samples, $L_j$ is the number of tokens in the $j$-th sample. The value of indicator function $\mathds{1}(g_{i,t}^{(\ell,j)} > 0)$ equals 1 if expert $i$ is selected (i.e., receives a non-zero gating weight), and 0 otherwise.
A higher $r_i^{(\ell)}$ indicates that expert $i$ is selected more frequently across samples of context-dependent tasks, reflecting its importance to context utilization.
Then, we identify the experts with top-$k$ high context-dependent ratio as context-faithful experts. 

\begin{table*}[t]
\small
\centering
\setlength{\tabcolsep}{9.2pt}{
\begin{tabular}{lcccccccccc}
\toprule
\multirow{2}{*}{\textbf{Models}} & \multicolumn{2}{c}{\textbf{SQuAD}} & \multicolumn{2}{c}{\textbf{NQ}} & \multicolumn{2}{c}{\textbf{HotpotQA}} & \multicolumn{2}{c}{\textbf{NQ-Swap}} & \multicolumn{2}{c}{\textbf{ConfiQA}}  \\
\cmidrule(lr){2-3}\cmidrule(lr){4-5}\cmidrule(lr){6-7}\cmidrule(lr){8-9}\cmidrule(lr){10-11}
~ & \textbf{EM} & \textbf{F1}  & \textbf{EM} & \textbf{F1} & \textbf{EM} & \textbf{F1} & \textbf{EM} & \textbf{F1} &\textbf{EM} & \textbf{F1} \\
\midrule
OLMoE-1B-7B
& 26.6 & 49.4
& 18.3 & 39.9
& 27.0 & 46.1
& 28.1 & 40.5
& 38.7 & 49.9 \\
\cellcolor{red!20}\quad\quad\quad  w/ Router Tuning
& \cellcolor{red!20}\textbf{80.5} & \cellcolor{red!20}\textbf{88.1} 
& \cellcolor{red!20}\textbf{62.4} & \cellcolor{red!20}\textbf{75.3} 
& \cellcolor{red!20}\textbf{60.4} & \cellcolor{red!20}\textbf{76.1} 
& \cellcolor{red!20}\textbf{76.4} & \cellcolor{red!20}\textbf{77.8} 
& \cellcolor{red!20}\textbf{76.9} & \cellcolor{red!20}\textbf{79.9} \\
\midrule
DeepSeek-V2-Lite
& 25.2 & 48.6
& 25.9 & 48.1
& 30.8 & 50.4
& 27.4 & 38.0
& 19.1 & 35.3 \\
\cellcolor{red!20}\quad\quad\quad  w/ Router Tuning
& \cellcolor{red!20}\textbf{83.6} & \cellcolor{red!20}\textbf{91.1}
& \cellcolor{red!20}\textbf{65.1} & \cellcolor{red!20}\textbf{77.5} 
& \cellcolor{red!20}\textbf{61.7} & \cellcolor{red!20}\textbf{77.7} 
& \cellcolor{red!20}\textbf{82.2} & \cellcolor{red!20}\textbf{84.3} 
& \cellcolor{red!20}\textbf{75.3} & \cellcolor{red!20}\textbf{78.2} \\
\midrule
MiniCPM-MoE-8x2B
& 45.8 & 65.1
& 35.1 & 55.8
& 38.9 & 57.3
& 42.8 & 50.5
& 38.6 & 48.2 \\
\cellcolor{red!20}\quad\quad\quad w/ Router Tuning
& \cellcolor{red!20}\textbf{80.5} & \cellcolor{red!20}\textbf{89.0} 
& \cellcolor{red!20}\textbf{61.1} & \cellcolor{red!20}\textbf{74.3} 
& \cellcolor{red!20}\textbf{60.7} & \cellcolor{red!20}\textbf{76.6} 
& \cellcolor{red!20}\textbf{71.7} & \cellcolor{red!20}\textbf{74.0} 
& \cellcolor{red!20}\textbf{65.7} & \cellcolor{red!20}\textbf{70.3} \\
\midrule
Mixtral-8x7B
& 21.1 & 43.5 
& 19.6 & 41.3 
& 25.5 & 43.4 
& 16.3 & 29.9 
& 12.3 & 20.3 \\
\cellcolor{red!20}\quad\quad\quad w/ Router Tuning
& \cellcolor{red!20}\textbf{49.6} & \cellcolor{red!20}\textbf{63.2} 
& \cellcolor{red!20}\textbf{44.0} & \cellcolor{red!20}\textbf{62.4} 
& \cellcolor{red!20}\textbf{57.0} & \cellcolor{red!20}\textbf{73.2} 
& \cellcolor{red!20}\textbf{64.1} & \cellcolor{red!20}\textbf{67.1} 
& \cellcolor{red!20}\textbf{67.6} & \cellcolor{red!20}\textbf{74.9} \\
\bottomrule
\end{tabular}}
\caption{The performance comparison between untuned and router tuned MoE models on context-dependent tasks.}
\label{tab:router_tuning_results}
\end{table*}

\subsection{Experimental Setting}
Below, we list the datasets, metrics, and models used in the empirical study. For implementation details, please refer to the Appendix \ref{app:hyperparameters}.

\paragraph{Dataset and Metrics}
We evaluate our approach on several widely used datasets for context-dependent tasks, including SQuAD \cite{DBLP:conf/emnlp/RajpurkarZLL16/squad}, NQ \cite{DBLP:journals/tacl/KwiatkowskiPRCP19/nq}, HotpotQA \cite{DBLP:conf/emnlp/Yang0ZBCSM18/hotpotqa}, NQ-Swap \cite{DBLP:conf/emnlp/LongprePCRD021/nq-swap}, and ConfiQA (Multi-Conflicts subset) \cite{DBLP:journals/corr/abs-2412-15280/confliqa}.
Dataset statistics are summarized in Table~\ref{tab:dataset_details}.
Among these, SQuAD, NQ, and HotpotQA are classic question answering benchmarks that span a spectrum of reasoning challenges upon context, from local comprehension to multi-hop reasoning.
Moreover, we also include two counterfactual QA benchmarks: NQ-Swap and ConfiQA, offering a more challenging and complementary evaluation of a model’s ability to identify and rely on the correct evidence.
For all datasets, we report Exact Match (EM) and token-level F1 scores as evaluation metrics. For additional dataset details, please refer to Appendix~\ref{app: dataset}.

\begin{table}[t]
\small
\centering
\setlength{\tabcolsep}{9pt}{
\begin{tabular}{lccc}
\toprule
\textbf{Models} & \textbf{Methods} & \textbf{EM} & \textbf{F1} \\
\midrule
\multirow{2}{*}{OLMoE-1B-7B} & Base & 24.1 & 34.6 \\
~ & \cellcolor{red!20}RT & \cellcolor{red!20}\textbf{100} & \cellcolor{red!20}\textbf{100} \\
\midrule
\multirow{2}{*}{MiniCPM-MoE-8x2B} & Base & 37.0 & 45.2 \\
~ & \cellcolor{red!20}RT & \cellcolor{red!20}\textbf{99.8} & \cellcolor{red!20}\textbf{99.8} \\
\bottomrule
\end{tabular}}
\caption{The performance of Router Tuning (RT) on CounterFact dataset that needs no complex reasoning.}
\label{tab:counterfact_results}
\end{table}

\paragraph{MoE Models}
We select four widely used open-source MoE-based LLMs for evaluation: OLMoE-1B-7B \cite{DBLP:journals/corr/abs-2409-02060/olmoe}, DeepSeek-V2-Lite \cite{DBLP:journals/corr/abs-2405-04434/deepseek-v2}, MiniCPM-MoE-8x2B \cite{DBLP:journals/corr/abs-2404-06395/minicpm}, and Mixtral-8x7B \cite{DBLP:journals/corr/abs-2401-04088/mixtral}.
These models cover a diverse range of configurations in terms of the number of experts (from 8 to 64) and overall model sizes (from 7B to 47B), providing a comprehensive testbed for analyzing expert behavior in context-dependent scenarios.
Detailed configurations for each model are listed in Table~\ref{tab:model_details}.
To ensure consistent and deterministic outputs across models, we adopt greedy decoding \cite{DBLP:conf/naacl/Germann03} for all experiments. For additional model details, please refer to Appendix~\ref{app: models}.

\begin{figure}[t]
\centering
\includegraphics[width=\columnwidth]{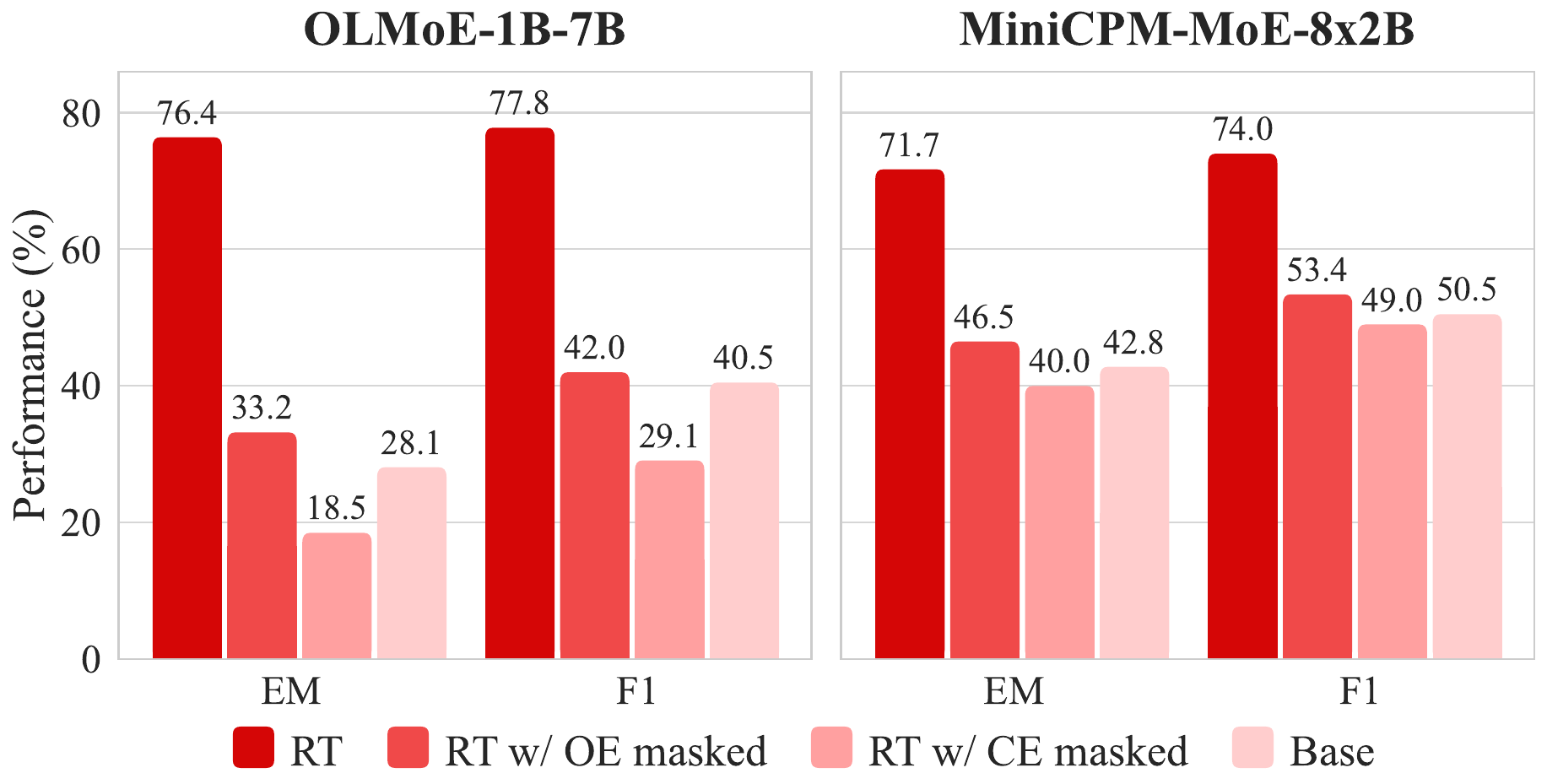}
\caption{Comparison of the performance impact on NQ-Swap when masking $k$ original experts (OE) vs. the top-$k$ context-faithful experts (CE), evaluated on router-tuned (RT) OLMoE-1B-7B and MiniCPM-MoE-8x2B models, relative to their respective base models (Base).}
\label{fig:block}
\end{figure}

\subsection{Results of Expert Elicitation}

In this section, we present the results of router tuning and analyze the role of context-faithful experts in enhancing context utilization.

Table~\ref{tab:router_tuning_results} summarizes the model performance on a variety of context-dependent benchmarks, both before and after router tuning. Across all evaluated models and tasks, router tuning consistently yields substantial improvements over the base models. This demonstrates that modifying only the expert selection mechanism significantly boosts the performance of context-dependent tasks, indicating the presence of \textbf{context-faithful experts}.

To assess whether the performance gains from the newly selected experts arise solely from enhanced compositional reasoning rather than effective context utilization, we also conduct experiments on the CounterFact dataset \cite{DBLP:conf/nips/MengBAB22}, 
where the model must rely on the provided counterfactual context to answer a simple question correctly. The task does not require complex reasoning, making it well-suited for isolating and evaluating the role of context-faithful experts.
Table \ref{tab:counterfact_results} reports the EM and F1 scores of OLMoE-1B-7B and MiniCPM-MoE-8x2B on CounterFact. We observe substantial performance improvements after router tuning, further providing strong evidence for the existence of context-faithful experts.

To examine the importance of these experts, we conduct a causal intervention experiment by masking the identified context-faithful experts on the router-tuned models. Specifically, by setting their gating weights to zero and measuring the resulting performance degradation. We conduct this experiment on the NQ-Swap dataset using the OLMoE-1B-7B and MiniCPM-MoE-8x2B models.
As illustrated in Figure~\ref{fig:block}, masking the context-faithful experts leads to substantial drops in performance, with EM score decreasing by 73.2\% for OLMoE-1B-7B and 44.2\% for MiniCPM-MoE-8x2B. In contrast, masking an equal number of originally selected experts results in smaller performance declines. These findings highlight that \textit{context-faithful experts play a critical role in solving context-dependent tasks}.

\begin{figure}[t]
\centering
\includegraphics[width=\columnwidth]{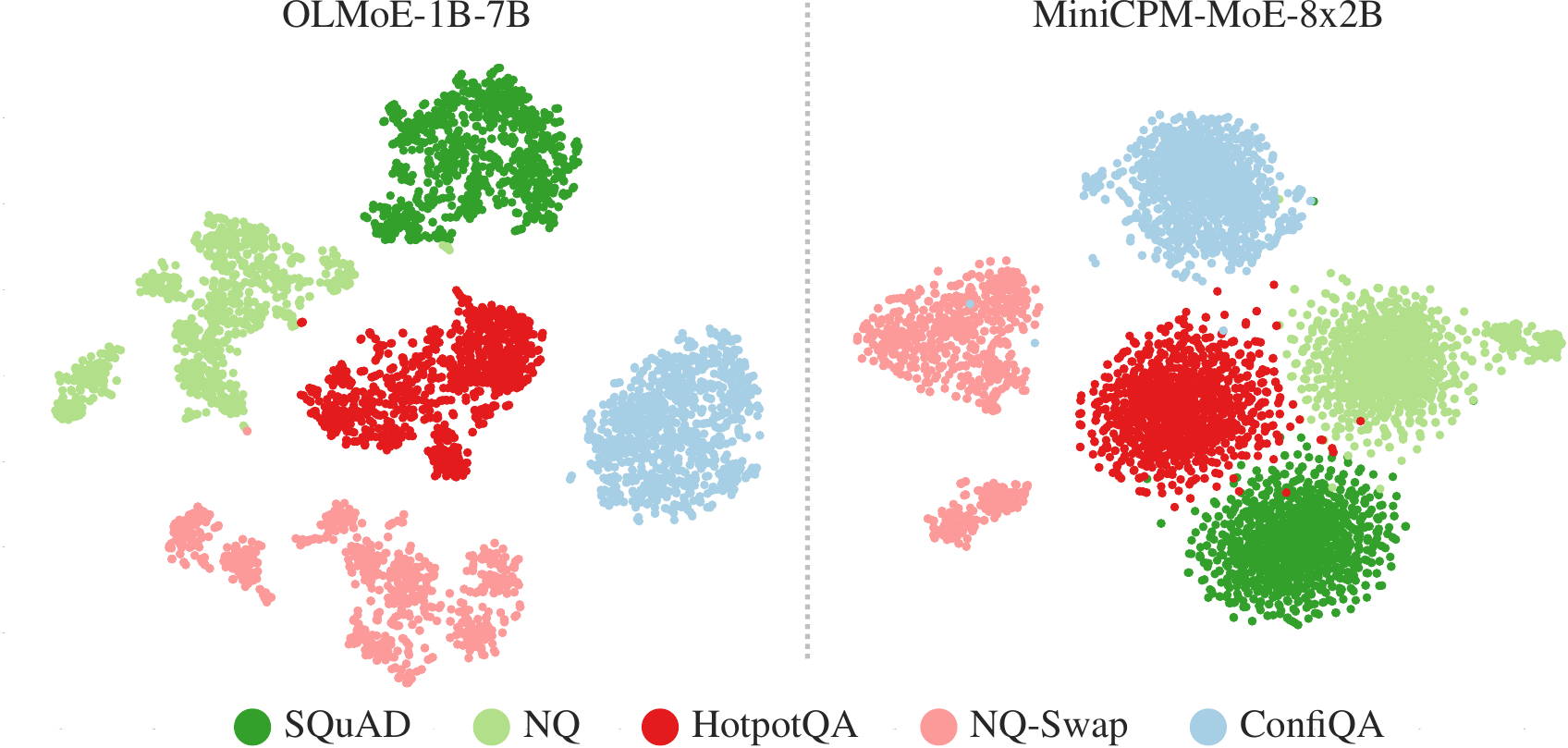}
\caption{t-SNE visualization of context-faithful expert activation patterns in OLMoE-1B-7B and MiniCPM-MoE-8×2B. For each model, 1,000 examples per dataset are randomly selected for projection.}
\label{fig:cluster}
\end{figure}

\subsection{Consistency of Context-Faithful Experts}

We further investigate whether the same context-faithful experts are consistently activated across different context-dependent tasks. To this end, we analyze expert activation patterns in OLMoE-1B-7B and MiniCPM-MoE-8×2B.
For each input sample, we calculate the activation frequency of context-faithful experts across all layers and concatenate these values into a feature vector. We then apply t-SNE \cite{van2008visualizing} to project these vectors into a 2D space for visualization.
As shown in Figure~\ref{fig:cluster}, samples from different datasets form clearly separable clusters in both OLMoE-1B-7B and MiniCPM-MoE-8×2B. This indicates that the router learns task-specific activation patterns for context-faithful experts, demonstrating that the model adapts its routing behavior based on the contextual requirements of each task, highlighting both the interpretability and task-awareness enabled by the router-tuning mechanism.

\begin{figure}[t]
\centering
\includegraphics[width=\columnwidth]{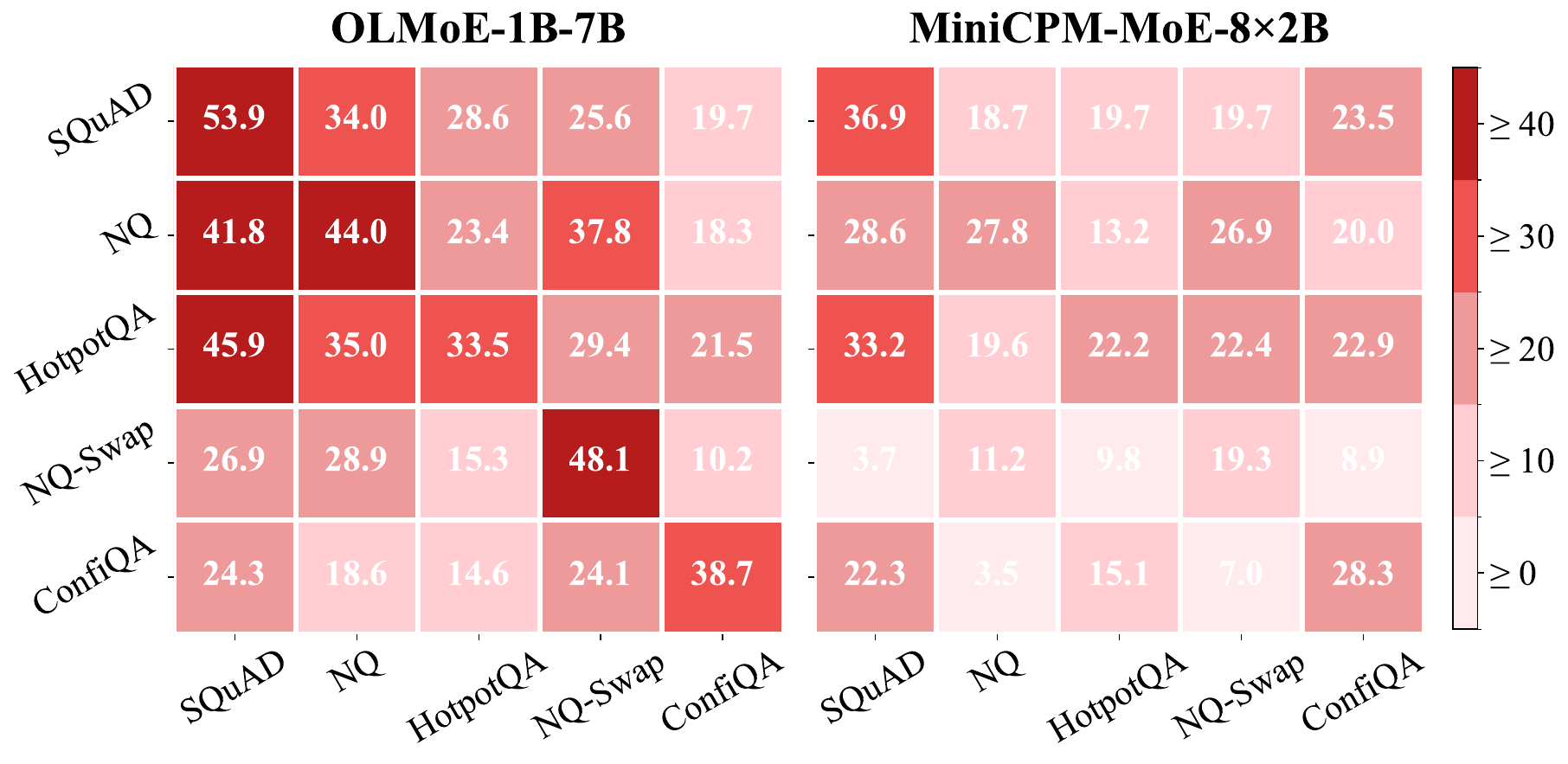}
\caption{Cross-task transfer performance of router-tuned models. Each cell shows the absolute EM score improvement over the base model, where the model is trained on the dataset in row $i$ and evaluated on the dataset in column $j$.}
\label{fig:transfer}
\end{figure}

While the results above suggest that different tasks benefit from distinct expert configurations, an important question remains: \textit{Can a tuned router network generalize its ability to activate context-faithful experts to unseen tasks?}
To explore this, we assess the transferability of the tuned router by applying model router-tuned on one dataset to other datasets, without any additional adaptation.
As shown in Figure~\ref{fig:transfer}, router-tuned models consistently outperform their base counterparts on unseen tasks. This suggests that the learned routing strategies capture generalizable and context-aware behaviors, enabling effective expert selection even outside the training domain.

\begin{figure*}[t]
  \centering
  \includegraphics[width=2.08\columnwidth ]{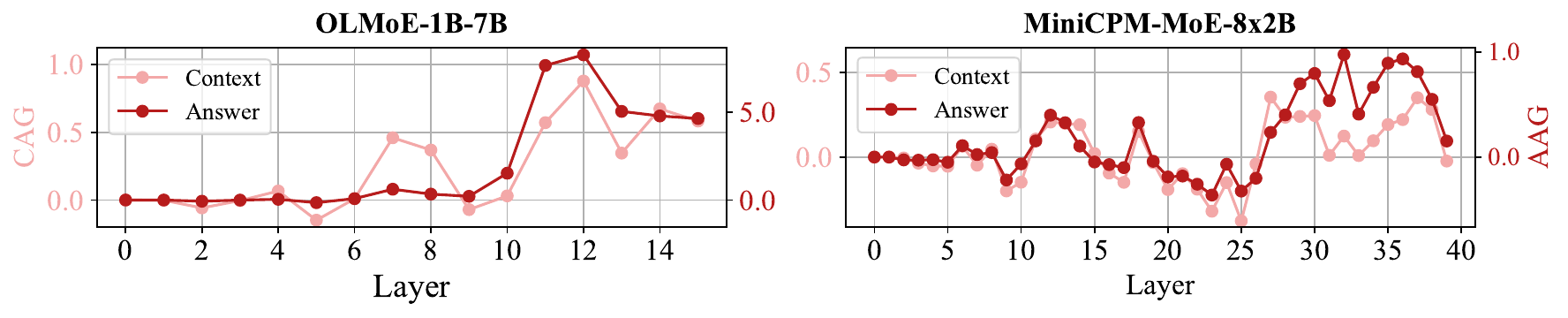}
  \caption{Visualization of layer-wise attention gain on context and answer (CAG and AAG) for the router-tuned model over the untuned model on the NQ-Swap test set. 
  }
  \label{fig:ATT_Comparation}
\end{figure*}

\subsection{Inner Working Mechanism of Context-faithful Experts}

Moreover, we investigate how context-faithful experts contribute to improving context faithfulness. In Transformer-based models, the self-attention mechanism plays a central role in perceiving and integrating contextual information \cite{DBLP:conf/iclr/SunZ0XZYSL25}. By assigning higher attention scores to relevant context tokens, the model can more effectively utilize the provided context.
To assess whether context-faithful experts enhance this mechanism, we examine whether their activation leads to increased attention over contextual tokens compared to the untuned model.
To this end, we introduce the \textbf{Context Attention Gain (CAG)} metric.

\begin{figure*}[t]
  \centering
  \subfloat[Layer 6]{\label{fig: olmoe_layer_var}\includegraphics[width=1.04\columnwidth]{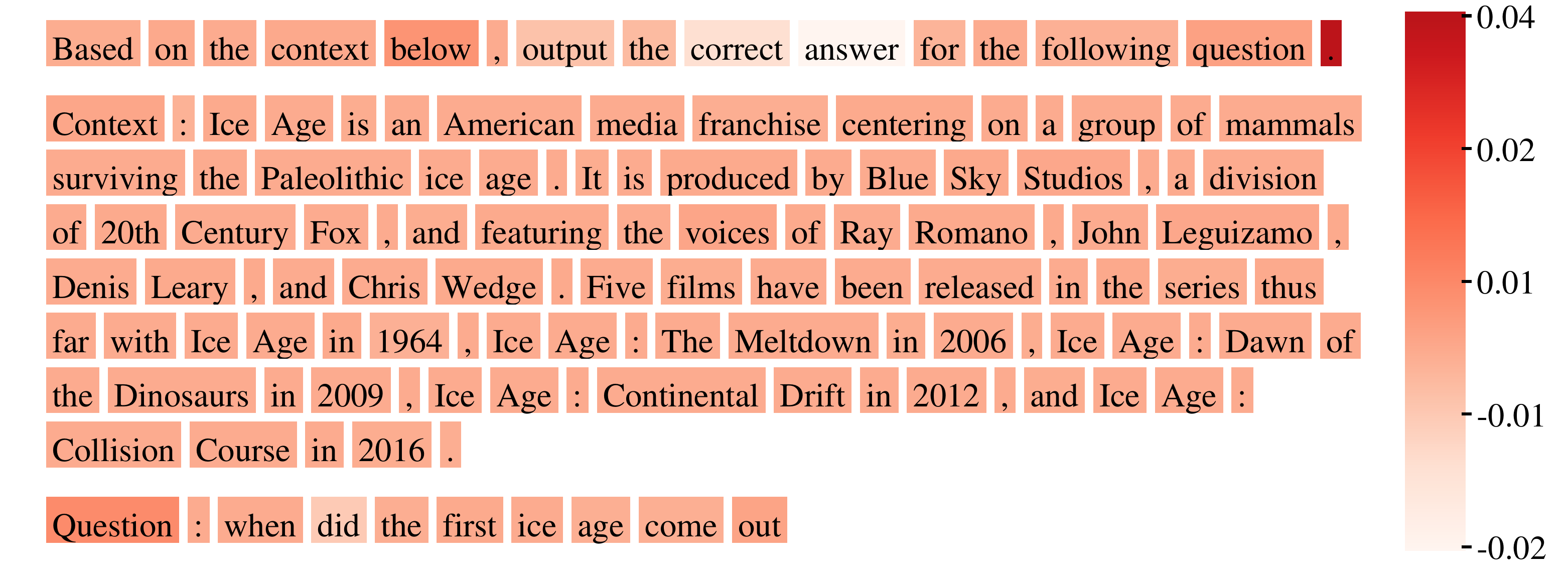}}
  \subfloat[Layer 12]{\label{fig: minicpm_layer_var}\includegraphics[width=1.04\columnwidth]{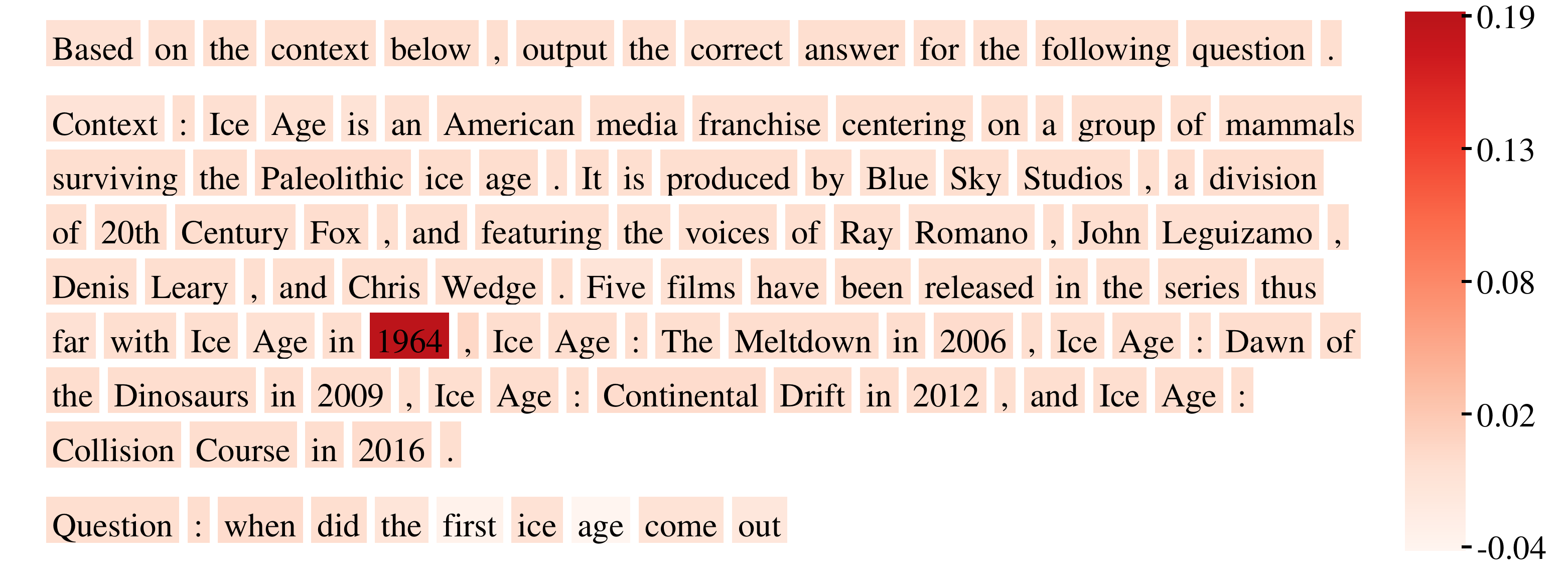}}
  \caption{Case study of attention gain from context-faithful experts in OLMoE-1B-7B on an NQ-Swap example. At (a) Layer 6 and (b) Layer 12, i.e., mid-level layer and deeper layer, the router-tuned model progressively increases attention to the context and answer tokens (i.e., ``1964''), illustrating a ``think twice'' mechanism. Notably, the base model fails on this example, while the router-tuned model provides the correct answer.}
  \label{fig:Case_ATT}
\end{figure*}

Let $\mathbf{A}_{i}^{(\ell)} \in \mathbb{R}^{N_h \times L_s \times L_s}$ denote the attention matrices at layer $\ell$ for the $i$-th sample, where $N_h$ is the number of heads and $L_s$ is the sequence length of input. Let $\mathcal{C}_i \subseteq \{1, \ldots, L_c\}$ represent the set of context token indices in sample $i$.
We average attention across all heads for the last token:
\begin{equation}
\bar{\mathbf{a}}_{i}^{(\ell)} = \frac{1}{N_h} \sum_{h=1}^{N_h} \mathbf{A}_{i}^{(\ell, h)}[L_s, :]
\end{equation}
Then, compute the attention mass assigned to all context tokens:
\begin{equation}
\alpha_{i}^{(\ell)} = \sum_{k \in \mathcal{C}_i} \bar{\mathbf{a}}_{i}^{(\ell)}[k]
\end{equation}
Let $\alpha_{i,\text{base}}^{(\ell)}$ and $\alpha_{i,\text{rt}}^{(\ell)}$ denote the attention to context tokens for the base and router-tuned models, respectively.
Finally, compute the average difference in context attention over all $N_s$ samples:
\begin{equation}
\text{CAG}^{(\ell)} = \frac{1}{N_s} \sum_{i=1}^{N_s} \frac{\alpha_{i,\text{rt}}^{(\ell)} - \alpha_{i,\text{base}}^{(\ell)}}{\alpha_{i,\text{base}}^{(\ell)}} 
\end{equation}
Similarly, we define \textbf{Answer Attention Gain (AAG)} indicator for analyzing the effect of context-faithful experts on the answer within context.

\begin{figure*}[t]
  \centering
  \includegraphics[width=2.06\columnwidth]{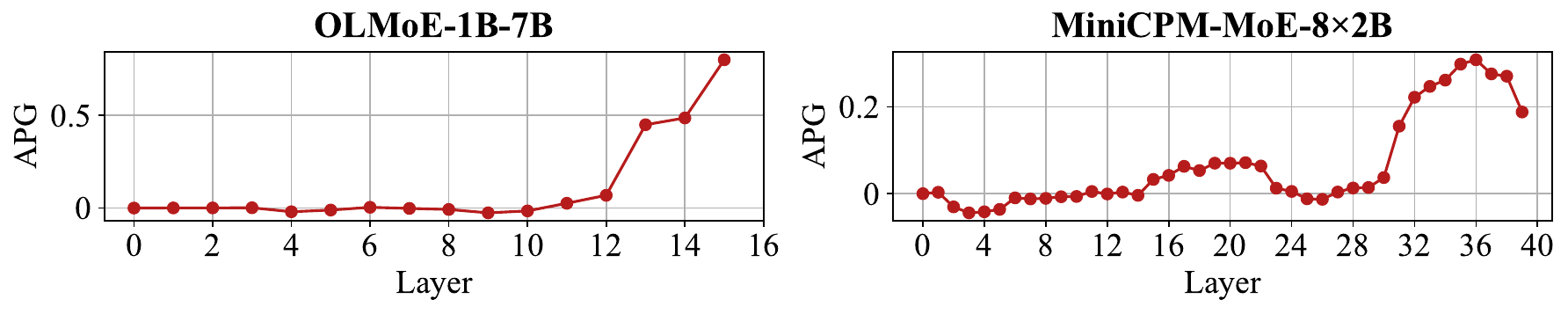}
  \caption{Layer-wise visualization of the Answer Probability Gain (APG) of the router-tuned models (OLMoE-1B-7B and MiniCPM-MoE-8x2B) over their untuned counterparts on the NQ-Swap test set, illustrating how context-faithful experts progressively enhance the model’s confidence in the correct answer across layers.}
  \label{fig:logit_lens}
\end{figure*} 

As shown in Figure~\ref{fig:ATT_Comparation}, we visualize the CAG and AAG scores across all layers for OLMoE-1B-7B-Instruct and MiniCPM-MoE-8×2B, evaluated on the NQ-Swap test set.
We observe that in both mid and deep layers, the router-tuned models allocate more attention to the overall context and, in particular, to the key answer tokens within the context, compared to their untuned counterparts. This amplification suggests that context-faithful experts effectively recalibrate the model’s focus toward the most relevant parts of the context.

We hypothesize that this layer-wise attention amplification reflects a two-stage reasoning process. As illustrated in Figure~\ref{fig:Case_ATT}, context-faithful experts in the mid-level layers help the model initially broaden its attention across the entire context—effectively “scanning” and identifying potentially relevant information. In the deeper layers, the model then narrows its focus, concentrating attention on the most critical spans within that context. In other words, the model appears to “think twice”: first by attending broadly to gather context, and then by selectively reinforcing attention on crucial details to inform its final decision 
(For more cases, please refer to Figure \ref{fig:More_Case_ATT}).

In addition, we define \textbf{Answer Probability Gain (APG)} metric to examine how context-faithful experts influence the model’s intermediate decision-making process. 
Let $\mathbf{h}_{i,last}^{(\ell)}\in\mathbb{R}^d$ be the hidden state output by the MoE module at layer $\ell$ for the last token of input $i$ where $d$ is hidden size, and let $y_i$ be its true answer tokens. 
We employ Logit Lens \cite{DBLP:conf/acl/DarG0B23/logitlens}, using the projection matrix $\mathbf{W}\in\mathbb{R}^{V\times d}$ (where $V$ is the vocabulary size) of language model head, to compute the probability $p_{m}^{(\ell)}(y_i)$ assigned to $y_i$ at layer $\ell$ under model $m\in\{\mathrm{base},\mathrm{rt}\}$ where $\mathrm{rt}$ refers to ``router-tuned''.
We then compute the average probability gain of the correct answer over all $N_s$ samples:
\begin{equation}
\text{APG}^{(\ell)}
= \frac{1}{N_s} \sum_{i=1}^{N_s}
\frac{p_{\mathrm{rt}}^{(\ell)}(y_i)
- p_{\mathrm{base}}^{(\ell)}(y_i)}{p_{\mathrm{base}}^{(\ell)}(y_i)}
\end{equation}
This metric quantifies how much context-faithful experts increase the model’s implicit confidence in the correct answer at each layer. 

Figure~\ref{fig:logit_lens} visualizes the layer-wise APG scores for OLMoE-1B-7B and MiniCPM-MoE-8×2B on the NQ-Swap test set.
We observe that, due to the context-faithful experts’ effective amplification of attention to both the broader context and the key answer tokens, the models are able to more accurately integrate relevant contextual information, leading to a substantial increase in the predicted probability of the correct answer in the deeper layers.

\section{Context Faithfulness Optimization}

\subsection{Context-faithful Expert Fine-Tuning}

Our previous analysis reveals a clear distinction between context-faithful experts and others: the former selectively amplify attention to relevant contextual information and key answer spans. By focusing on fine-tuning these high-impact experts, it is expected to achieve:  
(1) \textbf{Efficiency:} Significantly reduce the number of trainable parameters.  
(2) \textbf{Robustness:} Mitigate overfitting by preserving the pretrained weights of non-critical experts.  
(3) \textbf{Effectiveness:} Exploit the specialized routing behavior to further improve context utilization.

Motivated by this consideration, we propose \textbf{Context-faithful Expert Fine-Tuning (CEFT)} — a two-stage training strategy (Algorithm~\ref{alg:context_expert_tuning}). CEFT first identifies context-faithful experts using Router Lens analysis, 
then fine-tunes only these selected experts while freezing the rest of the model. 

\begin{table*}[t]
\small
\centering
\setlength{\tabcolsep}{11.8pt}{
\begin{tabular}{lcccccccccc}
\toprule
\multirow{2}{*}{\textbf{Methods}} & \multicolumn{2}{c}{\textbf{SQuAD}} & \multicolumn{2}{c}{\textbf{NQ}} & \multicolumn{2}{c}{\textbf{HotpotQA}} & \multicolumn{2}{c}{\textbf{NQ-Swap}} & \multicolumn{2}{c}{\textbf{ConfiQA}}  \\
\cmidrule(lr){2-3}\cmidrule(lr){4-5}\cmidrule(lr){6-7}\cmidrule(lr){8-9}\cmidrule(lr){10-11}
~ & \textbf{EM} & \textbf{F1}  & \textbf{EM} & \textbf{F1} & \textbf{EM} & \textbf{F1} & \textbf{EM} & \textbf{F1} &\textbf{EM} & \textbf{F1} \\
\midrule
\multicolumn{11}{c}{\textit{\textbf{OLMoE-1B-7B}}} \\
\midrule
FFT
& 81.6 & 88.8 
& 62.1 & 75.3 
& \underline{63.3} & \underline{78.8} 
& 88.3 & 88.8 
& 85.5 & 88.9  \\
ESFT
& \underline{82.6} & \underline{89.7} 
& \underline{62.6} & \underline{75.5} 
& 63.0 & 78.7 
& \textbf{91.4} & \textbf{91.7} 
& \underline{86.9} & \underline{89.2} \\
\cellcolor{red!20}CEFT
& \cellcolor{red!20}\textbf{83.1} 
& \cellcolor{red!20}\textbf{90.3} 
& \cellcolor{red!20}\textbf{64.1} 
& \cellcolor{red!20}\textbf{76.9} 
& \cellcolor{red!20}\textbf{63.8} 
& \cellcolor{red!20}\textbf{79.1} 
& \cellcolor{red!20}\underline{90.5} 
& \cellcolor{red!20}\underline{90.8} 
& \cellcolor{red!20}\textbf{87.1} 
& \cellcolor{red!20}\textbf{89.4} \\
\midrule
\multicolumn{11}{c}{\textit{\textbf{DeepSeek-V2-Lite}}} \\
\midrule
FFT
& 85.4 & 92.0 
& 67.7 & 79.6 
& 62.2 & 78.3 
& 91.3 & \underline{92.5} 
& \textbf{87.3} & \textbf{90.6}  \\
ESFT
& \underline{86.2} & \underline{92.9} 
& \underline{68.5} & \underline{80.3} 
& \textbf{65.9} & \textbf{81.6} 
& \underline{91.7} & \underline{92.5} 
& \underline{86.3} & 88.8 \\
\cellcolor{red!20}CEFT
& \cellcolor{red!20}\textbf{87.0} 
& \cellcolor{red!20}\textbf{93.5} 
& \cellcolor{red!20}\textbf{69.0} 
& \cellcolor{red!20}\textbf{81.2} 
& \cellcolor{red!20}\underline{63.7} 
& \cellcolor{red!20}\underline{80.1}
& \cellcolor{red!20}\textbf{92.7} 
& \cellcolor{red!20}\textbf{93.6} 
& \cellcolor{red!20}\textbf{87.3} 
& \cellcolor{red!20}\underline{89.4}  \\
\midrule
\multicolumn{11}{c}{\textit{\textbf{MiniCPM-MoE-8x2B}}} \\
\midrule
FFT
& 81.5 & 89.6 
& 62.8 & 75.9 
& 62.8 & 78.6 
& \underline{90.4} & \underline{91.0} 
& \textbf{86.3} & \textbf{88.8}  \\
ESFT
& \underline{83.3} & \underline{90.7} 
& \underline{65.9} & \underline{78.6} 
& \underline{64.3} & \underline{80.4} 
& 88.1 & 89.1 
& 84.4 & 87.2 \\
\cellcolor{red!20}CEFT
& \cellcolor{red!20}\textbf{83.8} 
& \cellcolor{red!20}\textbf{91.6} 
& \cellcolor{red!20}\textbf{66.6} 
& \cellcolor{red!20}\textbf{79.3} 
& \cellcolor{red!20}\textbf{64.9} 
& \cellcolor{red!20}\textbf{80.9}
& \cellcolor{red!20}\textbf{90.7} 
& \cellcolor{red!20}\textbf{92.0} 
& \cellcolor{red!20}\underline{84.8} 
& \cellcolor{red!20}\underline{87.5}  \\
\midrule
\multicolumn{11}{c}{\textit{\textbf{Mixtral-8x7B}}} \\
\midrule
FFT
& 65.3 & 76.5 
& 45.3 & 59.7 
& 56.3 & 73.6 
& \underline{68.0} & \underline{69.6} 
& \textbf{83.1} & \underline{87.8}  \\
ESFT
& \underline{66.1} & \underline{77.2} 
& \underline{45.9} & \underline{60.3} 
& \textbf{56.8} & \underline{73.7} 
& 67.2 & 68.4 
& 79.3 & 85.9 \\
\cellcolor{red!20}CEFT
& \cellcolor{red!20}\textbf{68.8} 
& \cellcolor{red!20}\textbf{78.9} 
& \cellcolor{red!20}\textbf{47.1} 
& \cellcolor{red!20}\textbf{61.1} 
& \cellcolor{red!20}\underline{56.7} 
& \cellcolor{red!20}\textbf{74.1} 
& \cellcolor{red!20}\textbf{68.7} 
& \cellcolor{red!20}\textbf{71.4} 
& \cellcolor{red!20}\underline{81.7} 
& \cellcolor{red!20}\textbf{87.9}  \\
\bottomrule
\end{tabular}}
\caption{Performance comparison of Fully Fine-Tuning (FFT), Expert-Specialized Fine-Tuning (ESFT) \citep{DBLP:conf/emnlp/WangCDXLW24/esft}, and Context-faithful Expert Fine-Tuning (CEFT). \textbf{Bold} numbers indicate the best performance, while \underline{underlined} numbers denote the second-best.}
\label{tab:ce_tuning_results}
\end{table*}

\begin{table}[t]
\small
\centering
\setlength{\tabcolsep}{7.5pt}{
\begin{tabular}{ccccc}
\toprule
\textbf{Training Dataset} & \textbf{Base} & \textbf{FFT} & \cellcolor{red!20}\textbf{CEFT} & \cellcolor{red!20}\textbf{RT} \\
\midrule
\multicolumn{5}{c}{\textit{\textbf{OLMoE-1B-7B}}} \\
\midrule
NQ-Swap   & 50.5 & 32.1 & \cellcolor{red!20}43.6 & \cellcolor{red!20}48.0  \\
ConfiQA   & 50.5 & 45.1 & \cellcolor{red!20}48.3 & \cellcolor{red!20}49.6  \\
\midrule
\multicolumn{5}{c}{\textit{\textbf{MiniCPM-MoE-8x2B}}} \\
\midrule
NQ-Swap   & 55.7 & 46.0 & \cellcolor{red!20}54.1 & \cellcolor{red!20}55.5  \\
ConfiQA   & 55.7 & 53.3 & \cellcolor{red!20}55.4 & \cellcolor{red!20}55.2  \\
\bottomrule
\end{tabular}}
\caption{Performance of OLMoE-1B-7B and MiniCPM-MoE-8x2B models on MMLU benchmark after employing different training methods.}
\label{tab:ood_performance}
\end{table}

\begin{algorithm}[t]
\caption{\textbf{CEFT}}
\label{alg:context_expert_tuning}
\begin{algorithmic}[1]
\Require Training dataset $\mathcal{D}_{\text{train}}$, MoE model $\mathcal{M}$, number of selected experts $k$
\Ensure Fine-tuned MoE model $\mathcal{M}^{*}$

\State \textbf{// Phase 1: Expert Identification}
\State Freeze all parameters of $\mathcal{M}$ except the router
\State Optimizing router parameters on $\mathcal{D}_{\text{train}}$

\For{each layer $\ell$ in the MoE model}
    \For{each expert $e_i$ in layer $\ell$}
        \State Compute context-dependence ratio $r_i^{(\ell)}$
    \EndFor
    \State Select top-$k$ experts with the highest $r_i^{(\ell)}$ as context-faithful experts $\mathcal{E}^{(\ell)}_{\text{context}}$
\EndFor

\State \textbf{// Phase 2: Identified Expert Fine-Tuning}
\State Freeze all parameters of $\mathcal{M}$ except the experts in $\mathcal{E}^{(\ell)}_{\text{context}}$ for each layer $\ell$
\State Train $\mathcal{M}$ on $\mathcal{D}_{\text{train}}$ to obtain final model $\mathcal{M}^{*}$

\end{algorithmic}
\end{algorithm}

\subsection{Empirical Results}

\paragraph{Main Performance}
To evaluate the effectiveness of the proposed CEFT approach, we compare it against two baseline strategies: standard \textbf{Fully Fine-Tuning (FFT)} and \textbf{Expert-Specialized Fine-Tuning (ESFT)} \cite{DBLP:conf/emnlp/WangCDXLW24/esft}. While ESFT also targets expert-level adaptation by only tuning a subset of experts deemed relevant to the task, it relies on the original, untuned router network to identify these experts, which may result in sub-optimal expert selection due to the influence of load balancing constraint and the limited task-awareness of the untuned router. 
In contrast, CEFT first adapts the router to reveal truly context-faithful experts—those that significantly enhance context-dependent reasoning—and then fine-tunes only these high-impact experts for more precise and efficient adaptation.
For implementation details, please refer to the Appendix \ref{app:hyperparameters}.
As shown in Table~\ref{tab:ce_tuning_results}, CEFT consistently matches or surpasses FFT and ESFT across MoE models and benchmarks, indicating its parameter-effectiveness in leveraging contextual information for improving generalization in context-dependent tasks.

\begin{figure}[t]
\centering
\includegraphics[width=\columnwidth]{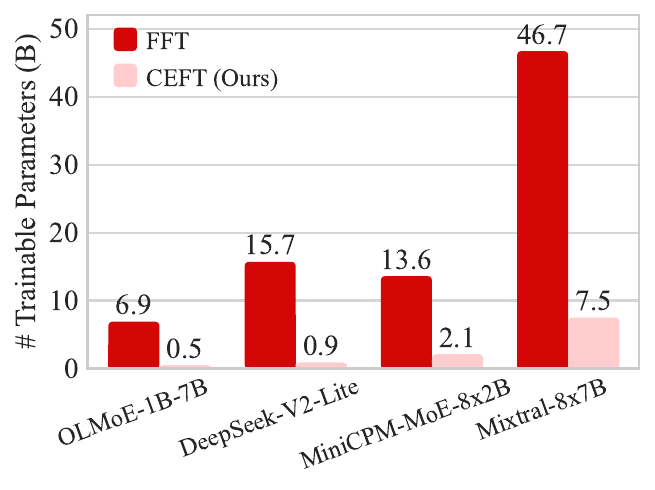}
\caption{Comparison of trainable parameters (fewer is better) between Fully Fine-Tuning (FFT) and Context-faithful Expert Fine-Tuning (CEFT).}
\label{fig:peft}
\end{figure}

\paragraph{Mitigation of Catastrophic Forgetting}
We further evaluate OLMoE-1B-7B and MiniCPM-MoE-8x2B on MMLU \cite{DBLP:conf/iclr/HendrycksBBZMSS21} to assess the impact of RT, FFT, and CEFT on the models’ original capabilities. As shown in Table \ref{tab:ood_performance}, the decline in MMLU performance is approximately proportional to the number of trainable parameters, with CEFT being substantially less susceptible to catastrophic forgetting compared to FFT.

\paragraph{Training Efficiency}
Figure~\ref{fig:peft} illustrates the substantial reduction in trainable parameters achieved by CEFT. For instance, OLMoE-1B-7B with FFT requires 6.9B parameters, whereas CEFT needs only 0.5B—a 13.8× reduction. This highlights CEFT as an efficient approach for adapting MoE models to context-dependent tasks, delivering comparable performance with far fewer trainable parameters than FFT.

\paragraph{Effect of Trainable Expert Count}
We conduct an ablation study on the number of training experts used in CEFT for OLMoE-1B-7B, using the NQ-Swap and ConfiQA datasets. Model performance is evaluated with varying numbers of experts: 1, 4, 8, 16, and 32.
As shown in the Table \ref{tab:num_exp_ablation_results}, increasing the number of experts generally improves performance up to a point, after which the gains plateau or slightly decline. This behavior can be attributed to the following: (1) Training too few experts may fail to capture enough context-faithful experts, limiting the model’s ability to leverage context information. (2) Training too many experts may involve irrelevant or noisy experts, leading to overfitting and degraded generalization.
These findings suggest that using a moderate number of experts provides a favorable trade-off between performance and efficiency in CEFT. Notably, we observe that setting the number of training experts to match the actual number of activated experts in the model often yields strong results (8 for OLMoE-1B-7B).

\begin{table}[t]
\small
\centering
\setlength{\tabcolsep}{12pt}{
\begin{tabular}{ccccc}
\toprule
\multirow{2}{*}{\textbf{\#Experts}} & \multicolumn{2}{c}{\textbf{NQ-Swap}} & \multicolumn{2}{c}{\textbf{ConfiQA}}  \\
\cmidrule(lr){2-3}\cmidrule(lr){4-5}
~ & \textbf{EM} & \textbf{F1}  & \textbf{EM} & \textbf{F1} \\
\midrule
1 & 87.5 & 88.5 & 84.4 & 88.0 \\
4 & 90.2 & 91.0 & 85.4 & 88.3 \\
8 & 90.5 & 90.8 & 87.1 & 89.4 \\
16 & 88.8 & 89.4 & 88.0 & 91.1 \\
32 & 88.1 & 88.8 & 85.6 & 88.8 \\
\bottomrule
\end{tabular}}
\caption{Performance of OLMoE-1B–7B model on NQ-Swap and ConfiQA with varying numbers of experts trained by CEFT.}
\label{tab:num_exp_ablation_results}
\end{table}

\section{Related works}

\paragraph{Context Faithfulness}
Context faithfulness refers to the extent to which a model's output remains accurate, consistent, and grounded in the provided context.
Unlike factual consistency, it emphasizes alignment specifically with the input context rather than with external world knowledge 
\cite{DBLP:conf/emnlp/ZhouZPC23,DBLP:conf/emnlp/XuQGW0ZX24}. 
In tasks such as RAG, maintaining context faithfulness is critical to ensure the reliability and trustworthiness of generated content \cite{DBLP:journals/corr/abs-2409-10102}. 
Recent studies have shown that LLMs often generate responses that are fluent yet contextually unfaithful \cite{DBLP:conf/acl/DuSSWSC24,DBLP:conf/iclr/Xie0CL024}. 
Various techniques, including prompt engineering \cite{DBLP:conf/emnlp/ZhouZPC23}, decoding constraints \cite{DBLP:conf/naacl/ShiHLTZY24}, attention amplification \cite{DBLP:conf/iclr/SunZ0XZYSL25}, mechanistic approaches \cite{DBLP:conf/emnlp/YuMP23,DBLP:conf/acl/OrtuJDSCS24,DBLP:conf/iclr/MinderDSMW0C25}, and learning-based methods such as RAG-fashioned fine-tuning \cite{DBLP:journals/corr/abs-2403-10131} and reinforcement learning with context-aware rewards \cite{DBLP:journals/corr/abs-2412-15280/confliqa}, have been proposed to improve grounding in the given context. 
Differently, our work focuses on an under-explored direction: leveraging the expert specialization in terms of the context utilization to improve the MoE performance on context-dependent tasks.

\paragraph{Expert Specialization}
The MoE architecture replaces the standard FFN layer \cite{DBLP:conf/nips/VaswaniSPUJGKP17/transformer_first} with a modular MoE layer comprising multiple parallel experts and a router network to sparsely activate top-$k$ experts \cite{DBLP:conf/icml/XueZFNZZ024/openmoe}.
During the pretraining stage of MoE, a load balancing loss is typically incorporated to force balanced expert utilization \cite{DBLP:conf/acl/DaiDZXGCLZYWXLH24/DeepSeekMOE}, which encourages the formation of expert specialization.
Recently, a lot of work has focused on the understanding of expert specialization, finding that different experts are responsible for different tokens \cite{DBLP:journals/corr/abs-2409-02060/olmoe}, domains \cite{DBLP:journals/corr/abs-2401-04088/mixtral}, tasks \cite{DBLP:conf/emnlp/WangCDXLW24/esft}, semantic units \cite{DBLP:conf/iclr/LiZ25}, or syntactic units \cite{DBLP:conf/coling/AntoineBL25}. These findings facilitate the targeted optimization of the specialized experts \cite{DBLP:conf/emnlp/WangCDXLW24/esft}. Inspired by these, our work investigates the expert specialization of context faithfulness.

\section{Conclusion}

In this work, we investigate expert specialization of context faithfulness in MoE models. We propose Router Lens to uncover context-faithful experts. Building upon this, our CEFT approach selectively updates the model parameters, yielding strong empirical gains while maintaining efficiency.

Looking forward, we see several exciting directions: combining mechanistic interpretability techniques such as SAE \cite{DBLP:conf/acl/KangWLWTWLWFZZ25} to further unravel MoE experts; extending our methods to other forms of specialization such as reflection \cite{DBLP:conf/acl/LiDLYWWZJZ25} and reasoning \cite{DBLP:journals/corr/abs-2506-08672}.
We hope this work inspires further research into aspect-specific expert discovery and optimization.


\section*{Limitations}
The proposed Router Lens relies on fine-tuning the router network to discover context-faithful experts.
Thus, it cannot identify such experts in a zero-shot or training-free setting. 
Moreover, improving the context utilization ability of these experts also depends on training procedures; we currently lack a method to directly enhance their behavior without additional supervision or optimization.

\section*{Ethics Statement} 
All experiments in this work are conducted on publicly available datasets commonly used in natural language processing research. No human subjects, personal data, or sensitive content are involved. We acknowledge that large-scale language models may carry risks of misuse, bias, or hallucination, and our work aims to improve their transparency and controllability, which may help mitigate such issues in the long term.

\section*{Acknowledgements}
This work was supported by the National Natural Science Foundation of China (No.62376031).

 
\bibliography{custom}

\begin{figure}[t]
\centering
\small
\begin{tcolorbox}[colback=gray!5!white, colframe=red!65!black, title=Prompt template used in experiments, width=\columnwidth]
\ttfamily
\textbf{Based on the context below, output the correct answer for the following question.}

\vspace{0.5em}
\textbf{Context}:
...Mission commander Paul and pilot Buzz Aldrin, landed the lunar module Eagle on July 20, 1969, at 20:18 UTC. Paul became the first human to step onto the lunar surface six hours after...

\vspace{0.5em}
\textbf{Question}:
Who took the first steps on the moon in 1969?
\end{tcolorbox}
\caption{Prompt template used during training and evaluation for context-dependent tasks.}
\label{fig:prompt-template}
\end{figure}

\appendix

\section{Implementation Details}
\label{app:hyperparameters}

\paragraph{Prompt Template}
To evaluate MoE models on context-dependent tasks, we adopt a unified prompt template as illustrated in Figure~\ref{fig:prompt-template}. This template explicitly instructs the model to generate an answer based on the given context and question.

\paragraph{Router Tuning}
We fine-tune the model for 1 epoch on SQuAD, NQ, and HotpotQA datasets, and 3 epochs on NQ-Swap and ConfiQA datasets. All experiments use the AdamW \cite{DBLP:conf/iclr/LoshchilovH19/adamw} optimizer with a learning rate of 5e-4, a warmup ratio of 0.1, and cosine learning rate decay. The batch size is set to 8, and the maximum sequence length is 300. All the experiments are run on one NVIDIA A100 80GB GPU.

\paragraph{Context-faithful Expert Fine-Tuning}
The hyperparameters are identical to those used in Router Tuning, except that CEFT fine-tunes the parameters of the context-faithful experts instead of the router.

\section{Datasets}
\label{app: dataset}
\paragraph{SQuAD} It is a widely used benchmark for extractive question answering \cite{DBLP:conf/emnlp/RajpurkarZLL16/squad}. Each example consists of a question and a context paragraph from Wikipedia, with the answer being a span within the paragraph (CC BY-SA 4.0 license).

\paragraph{NQ} It is a large-scale dataset collected from real user queries issued to the Google search engine \cite{DBLP:journals/tacl/KwiatkowskiPRCP19/nq}. Each question is paired with a Wikipedia page. Compared to SQuAD, NQ is more challenging due to its longer contexts and more diverse question types (Apache-2.0 license).

\paragraph{HotpotQA} It is a multi-hop question answering dataset where answering each question requires reasoning over multiple paragraphs including both supporting facts and distractor documents (CC BY-SA 4.0 license) \cite{DBLP:conf/emnlp/Yang0ZBCSM18/hotpotqa}.

\paragraph{NQ-Swap} It is a variant of the NQ \cite{DBLP:conf/emnlp/LongprePCRD021/nq-swap}, where part of the relevant document context is swapped with similar-looking but semantically misleading content, forcing the model to rely more heavily on context utilization (MIT License).

\paragraph{ConfiQA} It is designed to evaluate the context-faithfulness when knowledge conflicts \cite{DBLP:journals/corr/abs-2412-15280/confliqa}. The dataset is constructed by sampling real-world facts, generating multi-hop paths, and creating counterfactual contexts by replacing entities. We use its MC (Multi-Conflicts) subset that includes multiple counterfactuals (MIT License).

\begin{table}[t]
\small
\centering
\setlength{\tabcolsep}{12.8pt}{
\begin{tabular}{lccc}
\toprule
\textbf{Dataset} & \textbf{\#Train} & \textbf{\#Val} & \textbf{\#Test} \\
\midrule
SQuAD & 82,258 & 4,330 & 10,507 \\
NQ & 98,867 & 5,204 & 12,836 \\
HotpotQA & 69,281 & 3,647 & 5,904 \\
NQ-Swap & 3,000 & 746 & 1,000 \\
ConfiQA  & 4,000 & 500 & 1,500 \\
\bottomrule
\end{tabular}}
\caption{The statistics of datasets used in this work.}
\label{tab:dataset_details}
\end{table}

\begin{table}[t]
\small
\centering
\setlength{\tabcolsep}{3.8pt}{
\begin{tabular}{lcc}
\toprule
\textbf{Models} & \textbf{Params (A/F)} & \textbf{\#Experts (A/F)}  \\
\midrule
OLMoE-1B-7B & 1.3B / 6.9B & 8 / 64 \\
DeepSeek-V2-Lite & 2.4B / 15.7B & 6 / 64  \\
MiniCPM-MoE-8x2B & 4B / 13.6B & 2 / 8 \\
Mixtral-8x7B & 12.9B / 46.7B & 2 / 8  \\
\bottomrule
\end{tabular}}
\caption{Overview of the MoE models used in this work. A and F refer to ``Activated'' and ``Full''.}
\label{tab:model_details}
\end{table}

\section{Models}
\label{app: models}
\paragraph{OLMoE-1B-7B}
OLMoE-1B-7B is a fully open-source MoE developed by researchers from the Allen Institute for AI \cite{DBLP:journals/corr/abs-2409-02060/olmoe}. This model consists of 16 layers where 8 out of 64 experts are activated in each, and has 7 billion parameters but activates only 1 billion parameters per input token. 
We use its Instruct version in this work: \href{https://huggingface.co/allenai/OLMoE-1B-7B-0924-Instruct}{OLMoE-1B-7B-0924-Instruct}.

\paragraph{DeepSeek-V2-Lite}
This is a small version of the DeepSeek-V2 model \cite{DBLP:journals/corr/abs-2405-04434/deepseek-v2}. It has 27 layers where each MoE layer consists of 2 shared experts, 64 routed experts, and 6 activated experts. This configuration results in a total of 15.7 billion parameters, with 2.4 billion activated for each token. 
We use the parameters of its chat version: \href{https://huggingface.co/deepseek-ai/DeepSeek-V2-Lite-Chat}{DeepSeek-V2-Lite-Chat}.

\paragraph{MiniCPM-MoE-8x2B}
\href{https://huggingface.co/openbmb/MiniCPM-MoE-8x2B}{MiniCPM-MoE-8x2B} is a MoE variant of the MiniCPM model \cite{DBLP:journals/corr/abs-2404-06395/minicpm}. It initializes using sparse upcycling, replacing MLP layers with MoE layers. With two 2 of 8 experts activated, it results in approximately 4B activated parameters. This approach significantly boosts performance across various benchmarks, while maintaining computational efficiency.

\paragraph{Mixtral-8x7B}
Mixtral-8x7B is built upon the Mistral 7B model but incorporates 8 experts in each layer \cite{DBLP:journals/corr/abs-2401-04088/mixtral}. This design allows each token to access a vast pool of 47B parameters, yet only 13B parameters (2 experts in each layer) are actively used during inference.
We use its Instruct version, \href{https://huggingface.co/mistralai/Mixtral-8x7B-Instruct-v0.1}{Mixtral-8x7B-Instruct-v0.1}.

\begin{figure*}[t]
  \centering
  \subfloat[OLMoE-1B-7B]{\label{fig: olmoe_layer_var}\includegraphics[width=1.04\columnwidth]{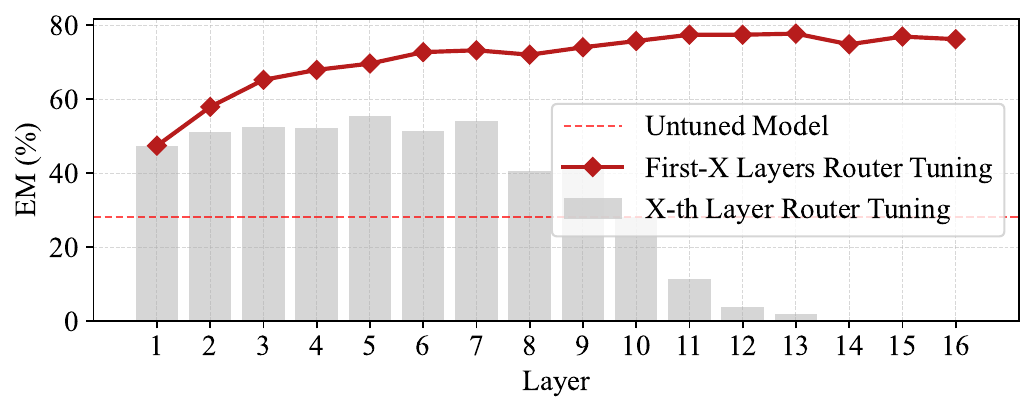}}
  \subfloat[MiniCPM-MoE-8x2B]{\label{fig: minicpm_layer_var}\includegraphics[width=1.04\columnwidth]{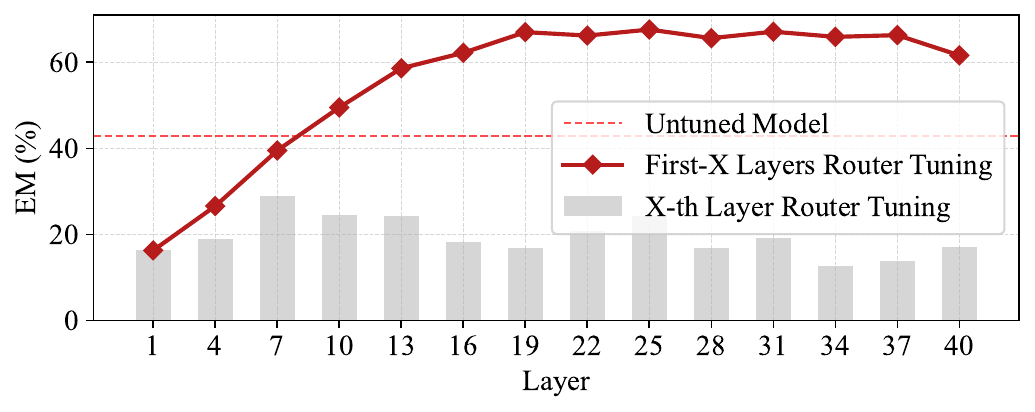}}
  \caption{The performance on NQ-Swap when fine-tuning the router of single layer (X-th Layer Router Tuning) and fine-tuning the routers of first layers (First-X Layers Router Tuning).}
  \label{fig:layer_var}
\end{figure*}

\begin{figure*}[t]
  \centering
  \subfloat[OLMoE-1B-7B]{\label{fig: olmoe_layer_var}\includegraphics[width=1.04\columnwidth]{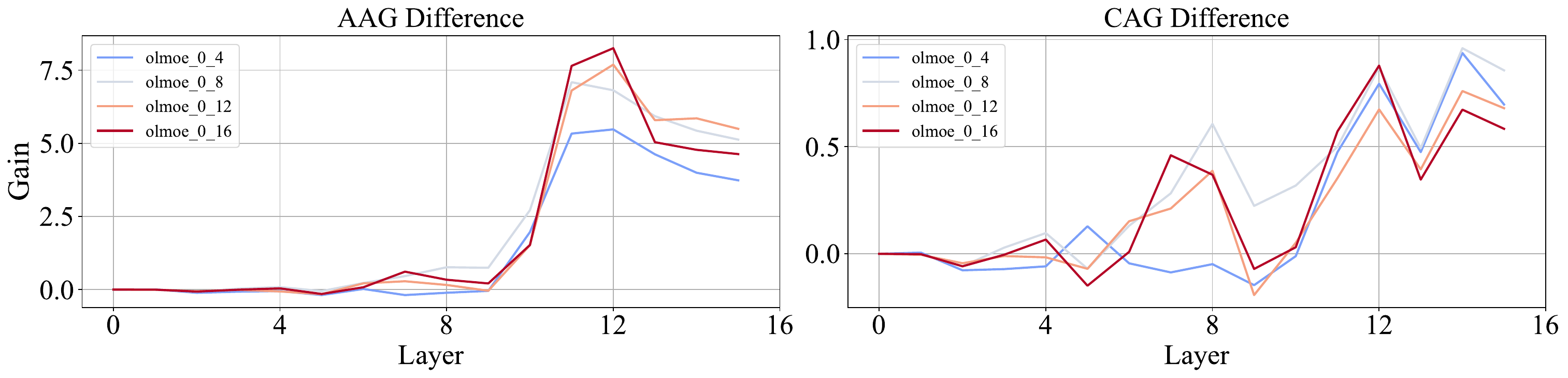}}
  \subfloat[MiniCPM-MoE-8x2B]{\label{fig: minicpm_layer_var}\includegraphics[width=1.04\columnwidth]{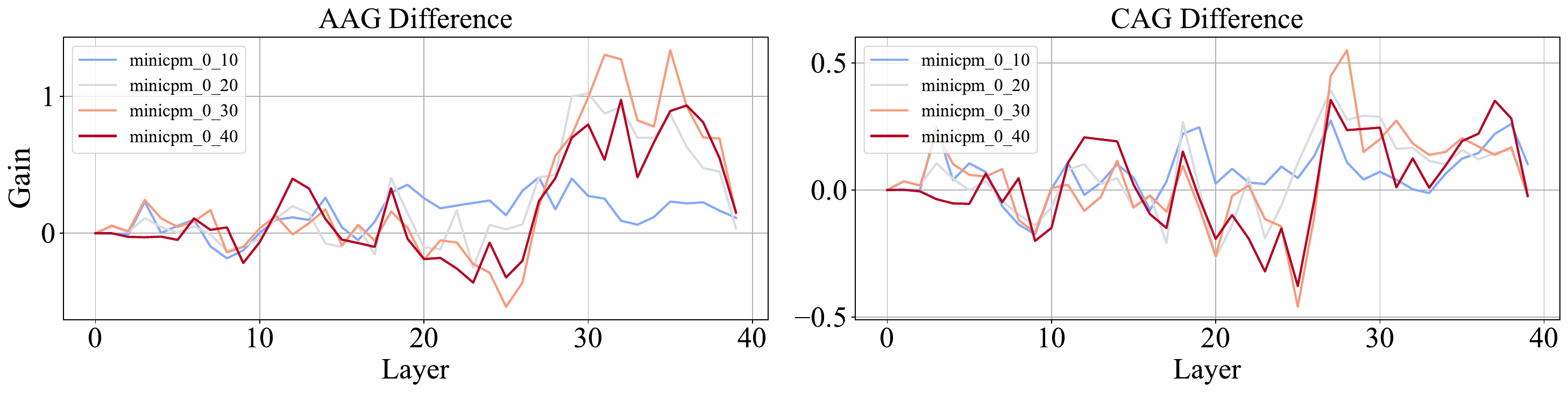}}
  \caption{Visualization of layer-wise Context Attention Gain (CAG) and Answer Attention Gain (AAG) for router-tuned models with varying numbers of tuned layers, relative to the untuned baseline, on the NQ-Swap test set.}
  \label{fig:0-X Finetune}
\end{figure*}

\begin{table}[t]
\small
\centering
\setlength{\tabcolsep}{4.7pt}{
\begin{tabular}{lcccccc}
\toprule
\multirow{2}{*}{\textbf{Methods}} & \multicolumn{3}{c}{\textbf{NQ-Swap}} & \multicolumn{3}{c}{\textbf{ConfiQA}}  \\
\cmidrule(lr){2-4}\cmidrule(lr){5-7}
~ & \textbf{Acc\textsubscript{LLM}} & \textbf{EM} & \textbf{F1} & \textbf{Acc\textsubscript{LLM}} & \textbf{EM} & \textbf{F1} \\
\midrule
FFT
& 91.9 & 88.3 
& 88.8 & 90.5
& 85.5 & 88.9 \\
ESFT
& 93.6 & 91.4 & 91.7 
& 90.8 & 86.9 & 89.2\\
\cellcolor{red!20}CEFT
& \cellcolor{red!20}92.4 & \cellcolor{red!20}90.5 & \cellcolor{red!20}90.8 
& \cellcolor{red!20}91.5 & \cellcolor{red!20}87.1 & \cellcolor{red!20}89.4 \\
\bottomrule
\end{tabular}}
\caption{The accuracy (Acc\textsubscript{LLM}) computed by LLM-as-a-judge, EM, and F1 scores of OLMoE-1B-7B on NQ-Swap and ConfiQA.}
\label{tab:llm_judge_results}
\end{table}

\section{Evaluation by LLM-as-a-judge}

EM and F1 are widely used metrics for evaluating question answering tasks. Both are standard in benchmarks such as SQuAD and NQ, ensuring comparability and providing a comprehensive assessment of model performance.
Additionally, we supplement our results with LLM-as-a-judge for OLMoE-1B-7B on the NQ-Swap and ConfiQA datasets. Specifically, we employ Qwen3-32B \cite{DBLP:journals/corr/abs-2505-09388} to assess the correctness of the MoE models’ predictions. 

The Table \ref{tab:llm_judge_results} reports the accuracy computed by the LLM-as-a-judge (Acc\textsubscript{LLM}), alongside the EM and F1 scores for OLMoE-1B-7B. As expected, Acc\textsubscript{LLM} tends to be higher than EM, reflecting its more permissive evaluation criteria. However, all three metrics show consistent performance trends across models and datasets.

\section{Layer-wise Context-faithful Expert Collaboration Analysis}
\label{app:expert_collaboration}
We investigate the collaborative effect of context-faithful experts—specifically, whether activating more such experts leads to greater performance gains on context-dependent tasks.

To this end, we perform two sets of fine-tuning experiments on the NQ-Swap dataset using OLMoE-1B-7B and MiniCPM-MoE-8×2B:
(1) X-th Layer Router Tuning, where we fine-tune the router of a single layer at a time, and
(2) First-X Layers Router Tuning, where we progressively fine-tune the routers of the first X layers.

As shown in Figure \ref{fig:layer_var}, the results reveal clear differences between two models. For X-th Layer Router Tuning, OLMoE-1B-7B shows substantial improvements even when tuning only a single early layer, suggesting that adjusting routing decisions at individual layers can significantly enhance context utilization. In contrast, MiniCPM-MoE-8×2B exhibits negligible gains from tuning single layer, implying that it requires multi-layer router adaptation to benefit from tuning. We hypothesize that this discrepancy arises partly from architectural differences: OLMoE-1B-7B employs more experts per layer (64) and activates more experts per token (8) than MiniCPM-MoE-8×2B, making each routing decision more influential and context-sensitive.
For First-X Layers Router Tuning, both models show incremental performance improvements as more layers are tuned, indicating that context-faithful experts across layers collaborate to improve performance. However, the gains gradually saturate with the inclusion of more tunable layers, suggesting that some experts across different layers may perform redundant roles.

\begin{figure}[t]
  \centering
  \includegraphics[width=1\columnwidth]{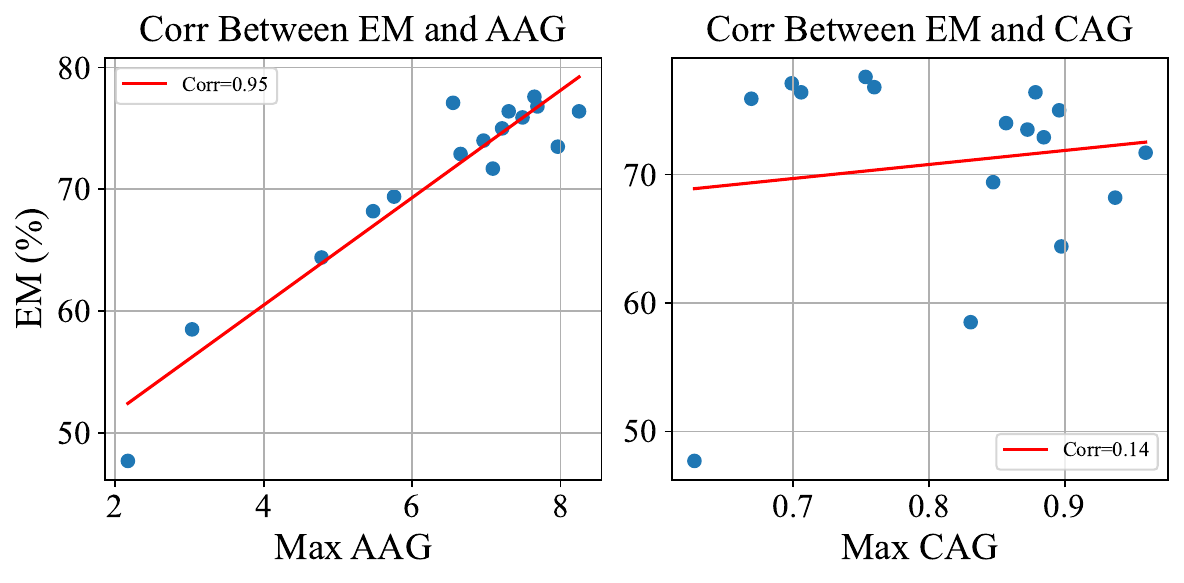}
  \caption{The correlation between the model's EM performance and the answer (context) attention gain.}
  \label{fig:EM-ATT}
\end{figure}

\begin{table}[t]
\small
\centering
\setlength{\tabcolsep}{9pt}{
\begin{tabular}{lccc}
\toprule
\textbf{Methods} & \textbf{BLEU} & \textbf{METEROR} & \textbf{ROUGE-L} \\
\midrule
\multicolumn{4}{c}{\textit{\textbf{OLMoE-1B-7B}}} \\
\midrule
Base & 1.9 & 35.5 & 21.7 \\
RT & 15.9 & 38.0 & 39.6 \\
FFT & 17.1 & 38.5 & 40.8 \\
\cellcolor{red!20}CEFT & \cellcolor{red!20}16.2 & \cellcolor{red!20}38.3 & \cellcolor{red!20}40.8 \\
\midrule
\multicolumn{4}{c}{\textit{\textbf{MiniCPM-MoE-8x2B}}} \\
\midrule
Base & 1.8 & 35.6 & 22.8 \\
RT & 11.6 & 34.8 & 36.8 \\
FFT & 15.6 & 38.7 & 40.3 \\
\cellcolor{red!20}CEFT & \cellcolor{red!20}15.9 & \cellcolor{red!20}38.3 & \cellcolor{red!20}39.7 \\
\bottomrule
\end{tabular}}
\caption{Performance on Gigaword benchmark.}
\label{tab:gigaword_results}
\end{table}

We further analyze the attention gain on both the context and the answer across different First-X Layers Router-Tuned models to examine whether different groups of context-faithful experts exhibit similar effects.
As shown in Figure \ref{fig:0-X Finetune}, the observed "think twice" phenomenon is not an isolated artifact caused by a particular combination of context-faithful experts. Instead, it is a consistent behavior that emerges across models with varying numbers of tuned layers.
To better understand the influence of this phenomenon—particularly the substantial increase in attention gain on both the context and the answer—we visualize the relationship via a scatter plot. Additionally, we compute the Pearson correlation coefficient between the attention gain on the answer and the corresponding performance improvements.
As illustrated in Figure \ref{fig:EM-ATT}, we observe a strong positive correlation (r = 0.95), indicating that greater attention gain on the answer is closely associated with higher model performance. This result provides further evidence that context-faithful experts enhance the model’s ability to focus on the most relevant contextual information in deeper layers, ultimately contributing to more context-faithful predictions.

\begin{figure*}[t]
  \centering
  \subfloat[Case 1-Layer 6]{\label{fig: olmoe_layer_var}\includegraphics[width=1.04\columnwidth]{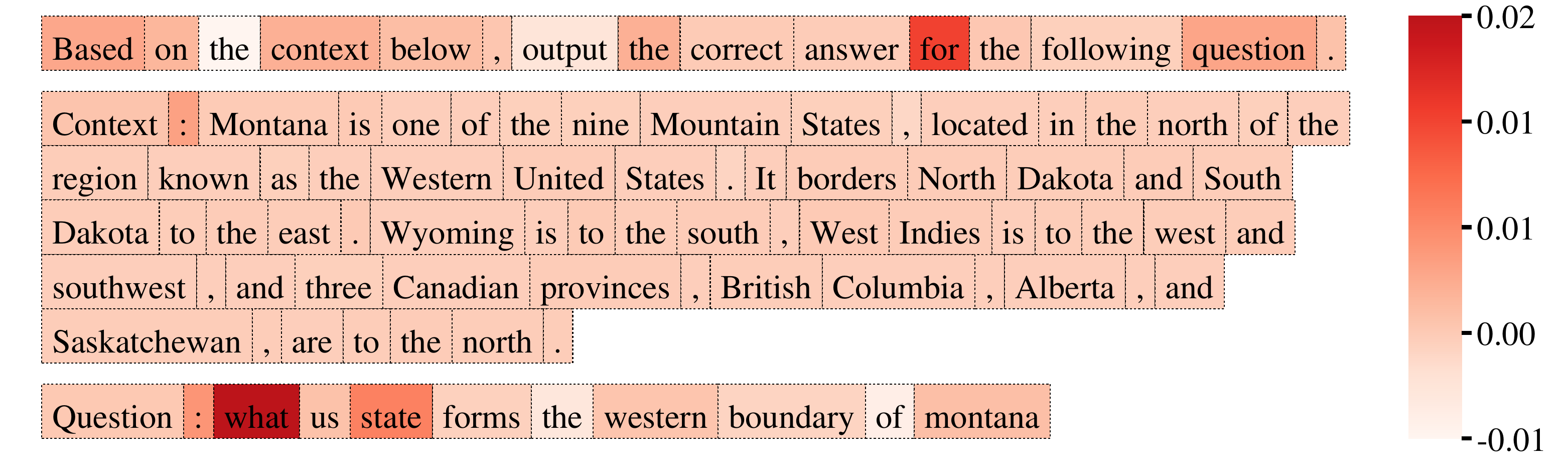}}
  \subfloat[Case 1-Layer 12]{\label{fig: minicpm_layer_var}\includegraphics[width=1.04\columnwidth]{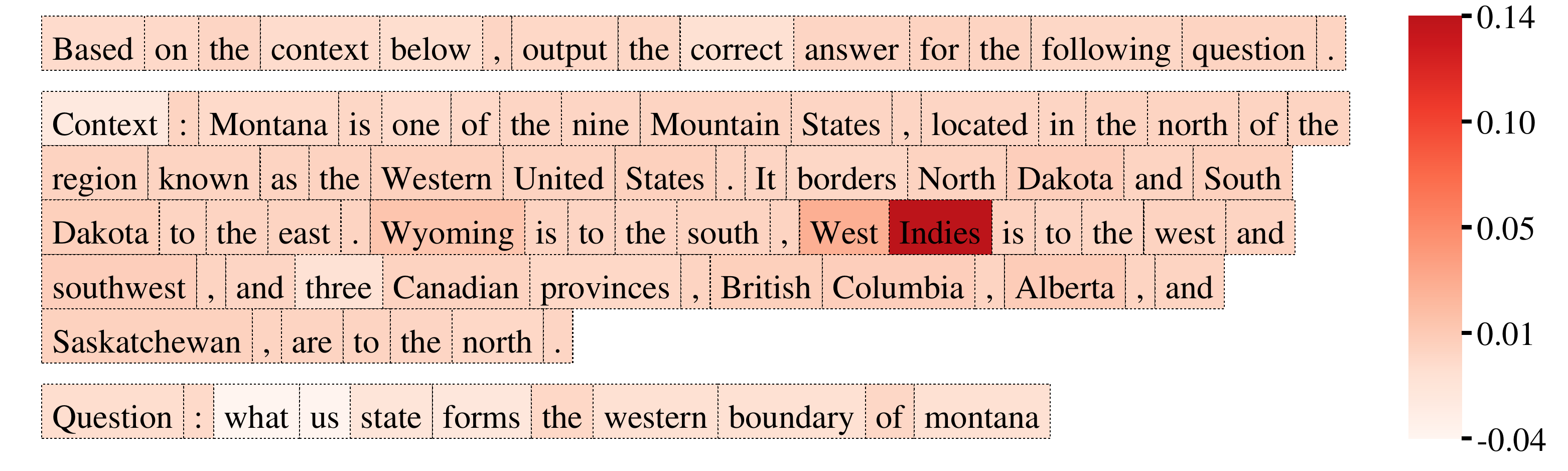}}
  \vspace{1em}
  \subfloat[Case 2-Layer 6]{\label{fig: olmoe_layer_var}\includegraphics[width=1.04\columnwidth]{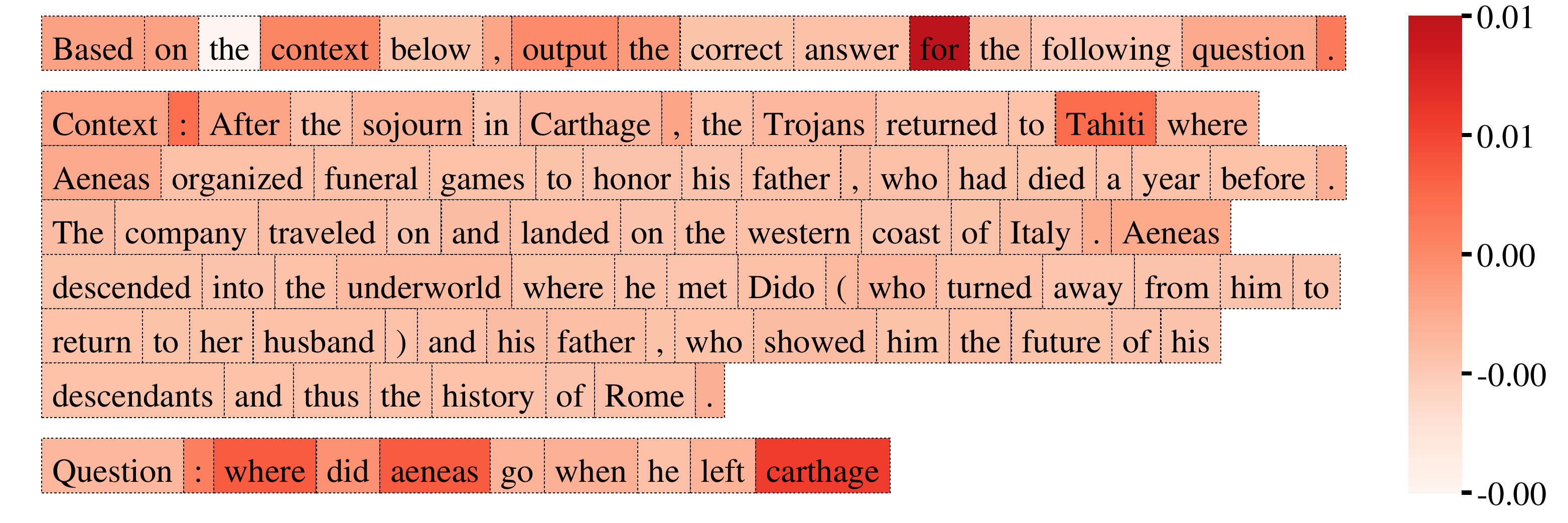}}
  \subfloat[Case 2-Layer 12]{\label{fig: minicpm_layer_var}\includegraphics[width=1.04\columnwidth]{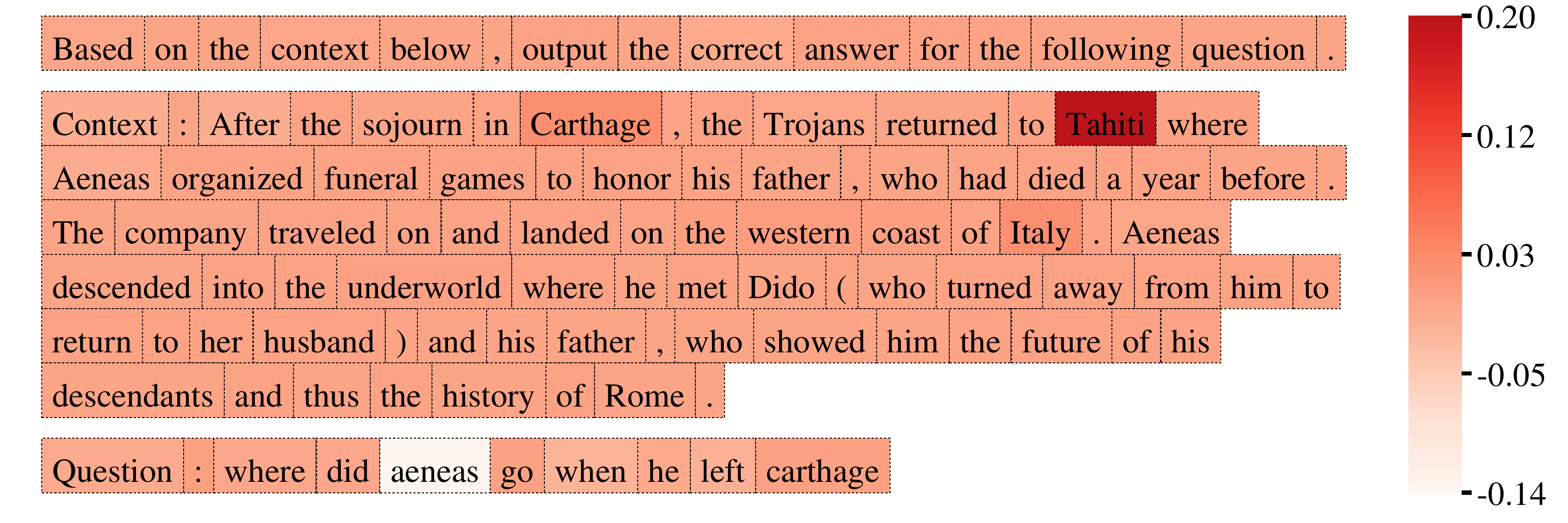}}
  \vspace{1em}
  \subfloat[Case 3-Layer 6]{\label{fig: olmoe_layer_var}\includegraphics[width=1.04\columnwidth]{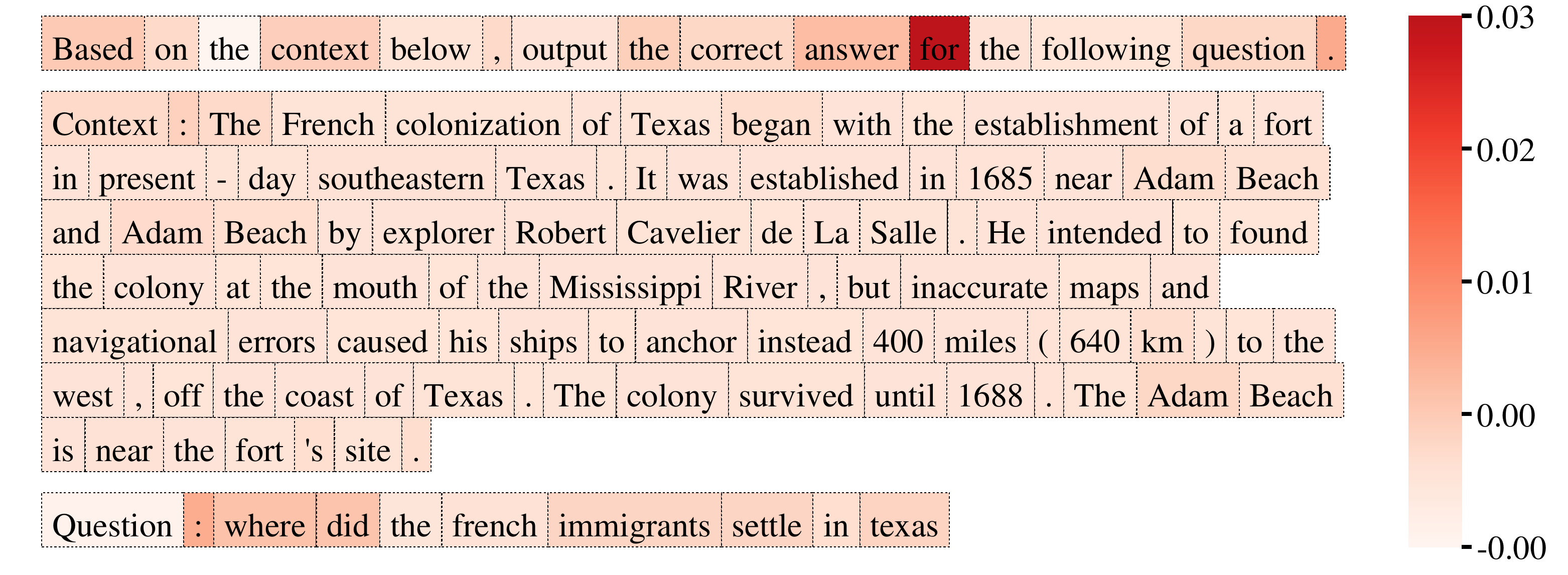}}
  \subfloat[Case 3-Layer 12]{\label{fig: minicpm_layer_var}\includegraphics[width=1.04\columnwidth]{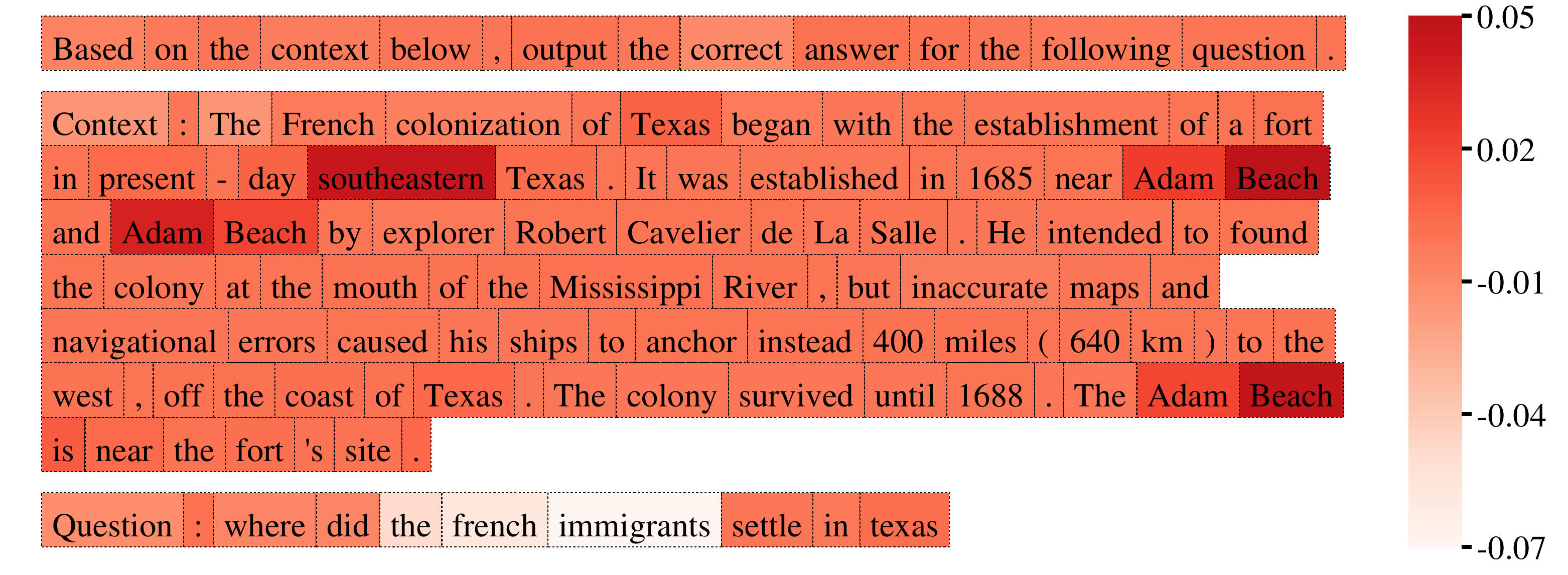}}
  \vspace{1em}
  \subfloat[Case 4-Layer 6]{\label{fig: olmoe_layer_var}\includegraphics[width=1.04\columnwidth]{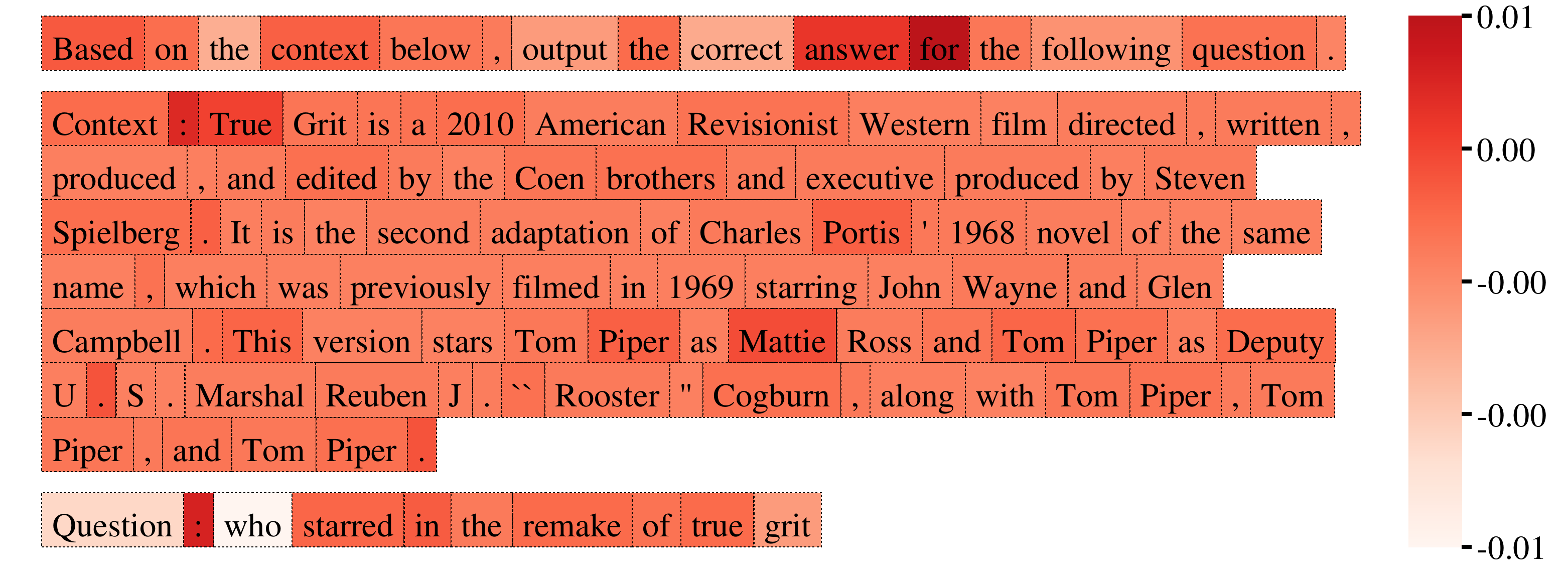}}
  \subfloat[Case 4-Layer 12]{\label{fig: minicpm_layer_var}\includegraphics[width=1.04\columnwidth]{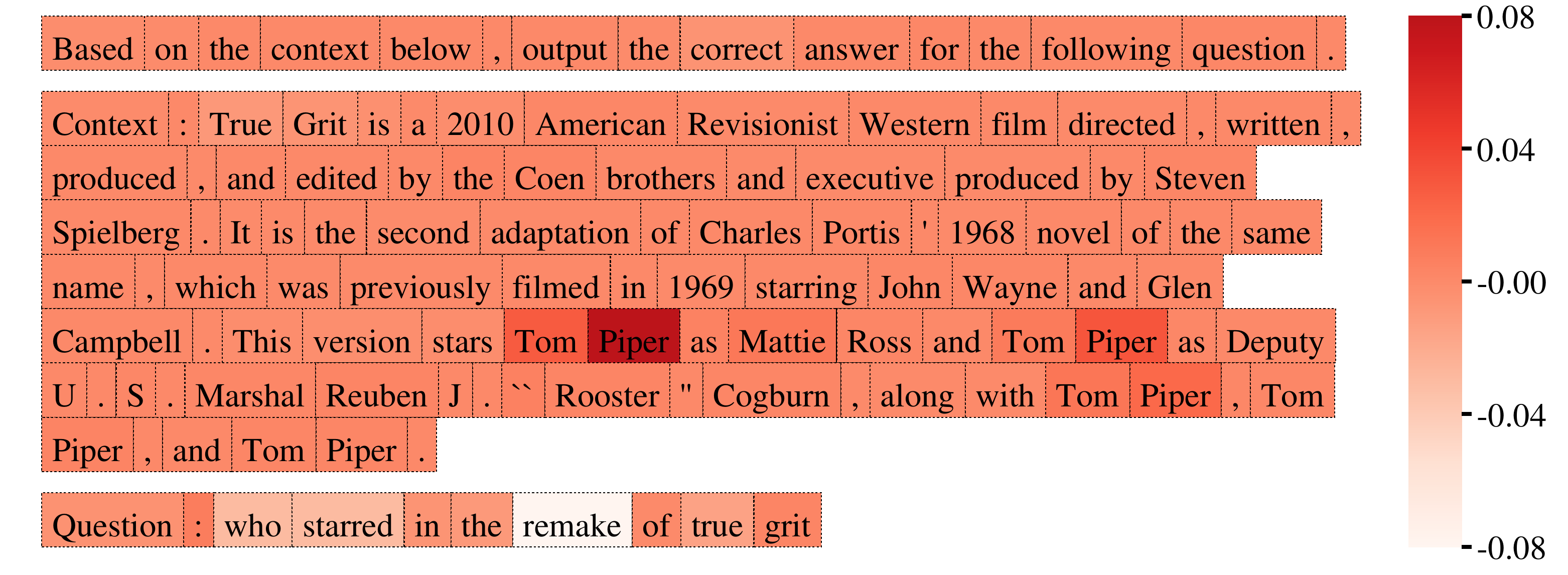}}
  \vspace{1em}
  \subfloat[Case 5-Layer 6]{\label{fig: olmoe_layer_var}\includegraphics[width=1.04\columnwidth]{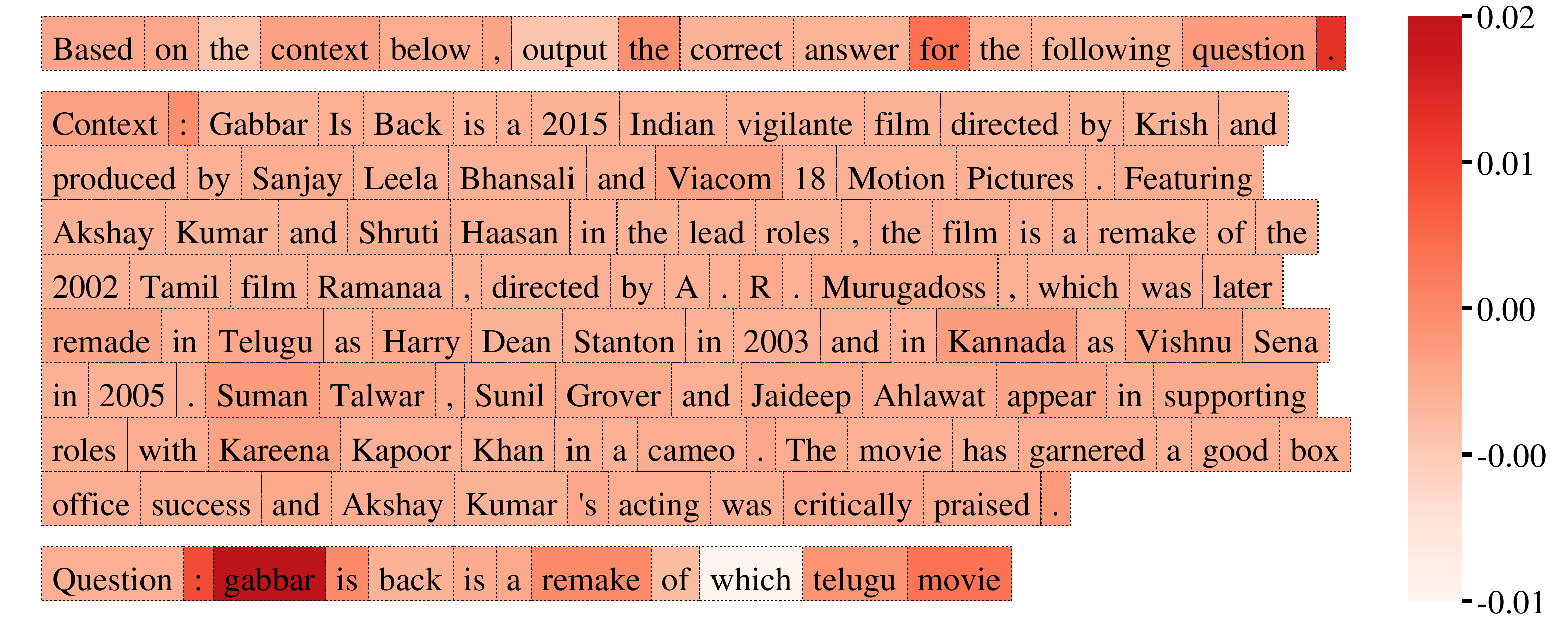}}
  \subfloat[Case 5-Layer 12]{\label{fig: minicpm_layer_var}\includegraphics[width=1.04\columnwidth]{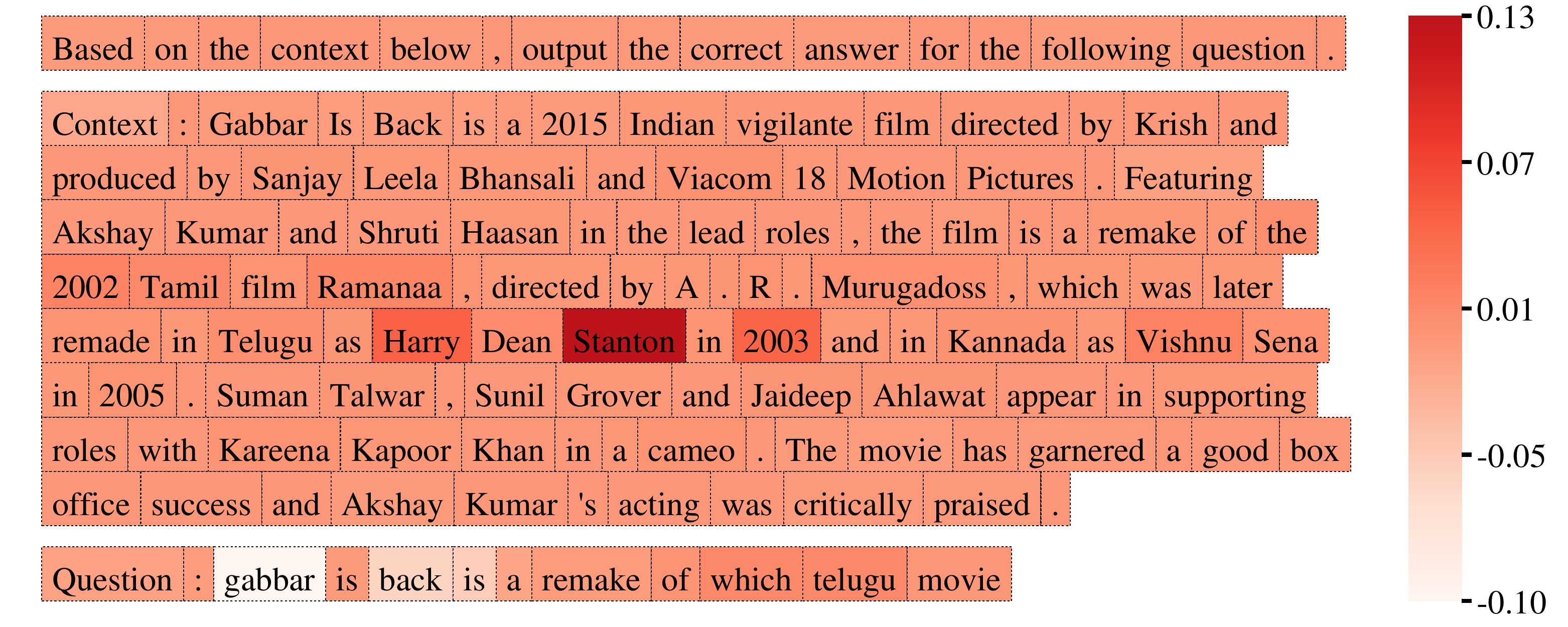}}
  \vspace{1em}
  \subfloat[Case 6-Layer 6]{\label{fig: olmoe_layer_var}\includegraphics[width=1.04\columnwidth]{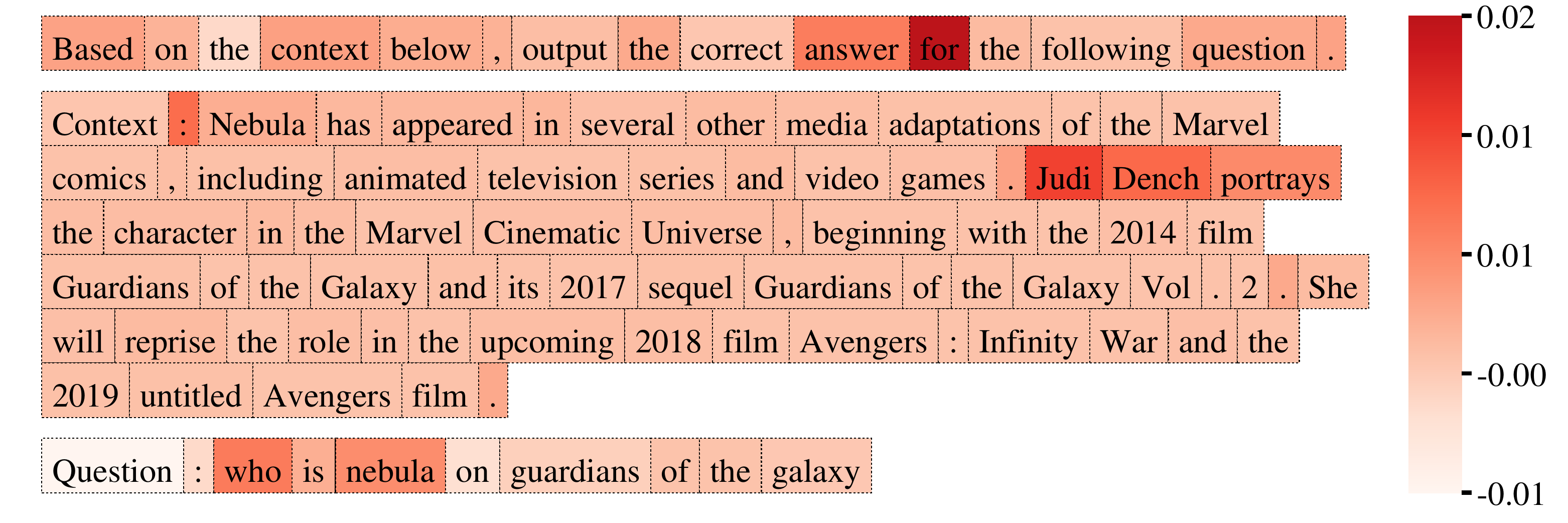}}
  \subfloat[Case 6-Layer 12]{\label{fig: minicpm_layer_var}\includegraphics[width=1.04\columnwidth]{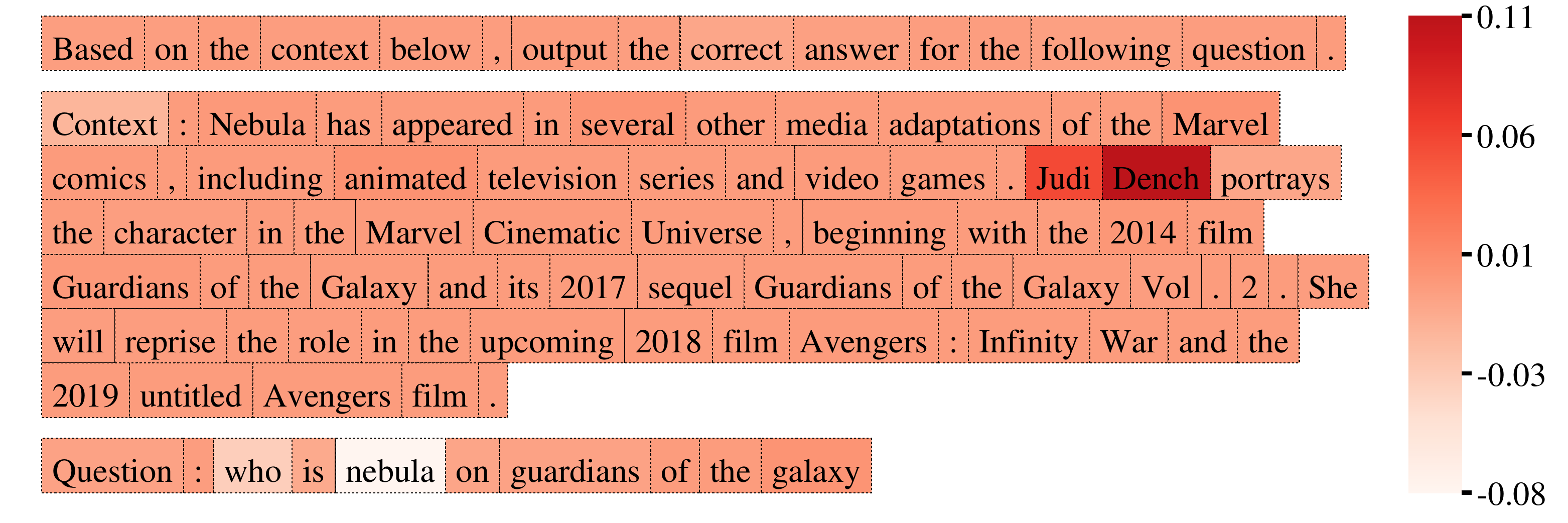}}
    \caption{More cases of attention gain from context-faithful experts in OLMoE-1B-7B on NQ-Swap examples. The correct answers to these examples are ``West Indies'', ``Tahiti'', ``Adam Beach'', ``Tom Piper'', ``Harry Dean Stanton'', and ``Judi Dench'', respectively.}
  \label{fig:More_Case_ATT}
\end{figure*}

\section{Performance on Non-QA Task}
We also examine RT and CEFT on other context-dependent task, e.g., summarization. We evaluate OLMoE-1B-7B and MiniCPM-MoE-8x2B on the widely used Gigaword benchmark \cite{DBLP:conf/emnlp/RushCW15}, where we sample 2000/500/500 examples as the train/validation/test sets, respectively.

Table \ref{tab:gigaword_results} reports their performance in terms of BLEU, METEOR, and ROUGE-L scores.
Our experiments show that router tuning consistently improves summarization performance over the base model. Furthermore, with CEFT—which selectively tunes the experts identified by the tuned router—we achieve performance comparable to full fine-tuning.
These results validate the effectiveness and generalizability of our method beyond contextual QA and reasoning tasks.

\section{Performance on Task Independent on the Context}

We further evaluate CEFT on the MemoTrap dataset \cite{DBLP:conf/naacl/ShiHLTZY24}, a benchmark specifically designed to detect whether language models memorize and regurgitate training data. Notably, MemoTrap is independent of any additional context, and therefore does not require context faithfulness.

Table \ref{tab:memotrap_results} reports the accuracy of OLMoE-1B-7B and MiniCPM-MoE-8x2B on MemoTrap. From these results, we observe the following in this context-independent task:
(1) Router tuning helps the model activate the appropriate experts for the task and leads to significant performance improvement.
(2) CEFT, by selectively training the most relevant experts, achieves performance comparable to full fine-tuning.
These findings demonstrate the effectiveness of both Router Lens and CEFT beyond context-dependent tasks. 

\begin{table}[t]
\small
\centering
\setlength{\tabcolsep}{6pt}{
\begin{tabular}{lcccc}
\toprule
\textbf{Models} & \textbf{Base} & \textbf{RT} & \textbf{FFT} & \cellcolor{red!20}\textbf{CEFT} \\
\midrule
OLMoE-1B-7B & 56.6 & 89.0 & 96.5 & \cellcolor{red!20}96.1 \\
MiniCPM-MoE-8x2B & 63.5 & 86.4 & 96.5 & \cellcolor{red!20}95.1 \\
\bottomrule
\end{tabular}}
\caption{Performance on MemoTrap dataset.}
\label{tab:memotrap_results}
\end{table}

\begin{table}[t]
\small
\centering
\setlength{\tabcolsep}{19.5pt}{
\begin{tabular}{lcc}
\toprule
\textbf{Methods} & \textbf{EM} & \textbf{F1} \\
\midrule
Base & 28.1 & 40.5 \\
ContextCite & 28.9 & 41.2 \\
CFP & 46.1 & 51.6 \\
CAD & 55.9 & 60.8 \\
Context-DPO & 65.7 & 69.8 \\
\midrule
\cellcolor{red!20}CEFT & \cellcolor{red!20}90.5 & \cellcolor{red!20}90.8 \\
\quad w/ ContextCite & 90.7 & 91.2 \\
\quad w/ CAD & 90.8 & 91.0 \\
\bottomrule
\end{tabular}}
\caption{Performance comparison of methods designed for improving context faithfulness on NQ-Swap.}
\label{tab:other_results}
\end{table}

\section{Comparison with Other Methods for Context Utilization}

We further compare our CEFT approach with Context-Faithful Prompting (CFP) \cite{DBLP:conf/emnlp/ZhouZPC23}, Context-Aware Decoding (CAD) \cite{DBLP:conf/naacl/ShiHLTZY24}, ContextCite \cite{DBLP:conf/nips/Cohen-WangSGM24}, and Context DPO \cite{DBLP:journals/corr/abs-2412-15280/confliqa} on the NQ-Swap and ConfiQA datasets. Among these, CFP, CAD, and ContextCite are unsupervised methods optimizing for prompt engineering, decoding, and context pruning, respectively. 
In contrast, Context DPO adopts a direct preference optimization to guide LLMs to prefer context-faithful outputs.

Table \ref{tab:other_results} shows the performance comparison between CEFT and the above context-utilization methods. Across both datasets—NQ-Swap and ConfiQA, CEFT consistently achieves the best results, significantly outperforming all baselines in both EM and F1 scores.
These results highlight the effectiveness of CEFT in leveraging context-faithful experts to model context-sensitive behavior more accurately than prior unsupervised prompting or decoding strategies, as well as preference-optimized approaches.

We combine ContextCite and CAD with CEFT to explore potential improvements, respectively. However, neither ContextCite nor CAD yields substantial gains. A possible explanation is that both context-level and decoding-level optimizations have limited capacity to enhance performance in this setting, particularly since they are unsupervised and not directly aligned with the model’s training objectives.

\end{document}